\documentclass[lettersize,journal]{IEEEtran}
\usepackage{amsmath,amsfonts}
\usepackage{algorithmic}
\usepackage{array}
\usepackage{textcomp}
\usepackage{stfloats}
\usepackage{verbatim}
\usepackage{graphicx}
\usepackage{balance}
\usepackage{amssymb}
\usepackage{epsfig}
\usepackage{multirow}
\usepackage{tabularx}
\usepackage{adjustbox}
\usepackage{siunitx}
\usepackage{subfig}

\usepackage{stmaryrd} 
\usepackage{latexsym} 
\usepackage{url}

\usepackage[breaklinks,colorlinks,bookmarks]{hyperref}
\usepackage{booktabs}

\usepackage{xcolor,colortbl}
\definecolor{Gray1}{gray}{0.85}
\definecolor{Gray2}{gray}{0.65}
\definecolor{darkgreen}{RGB}{0,127,0}
\definecolor{darkred}{RGB}{200,0,0}

\usepackage{pifont}
\newcommand{\cmark}{\ding{51}}%
\newcommand{\xmark}{\ding{53}}%

\usepackage[capitalize]{cleveref}
\crefname{section}{Sec.}{Secs.}
\Crefname{section}{Section}{Sections}
\Crefname{table}{Table}{Tables}
\crefname{table}{Tab.}{Tabs.}

\newcommand{\ie}{\textit{i.e.}}
\newcommand{\eg}{\textit{e.g.}}

\begin{document}

\title{Deep Depth Estimation from Thermal Image \\: Dataset, Benchmark, and Challenges}

\author{Ukcheol Shin, Jinsun Park
\thanks{Manuscript received March XX, 2025; revised April XX, 2025.}
%
%
\thanks{This work was supported by the National Research Foundation of Korea(NRF) grant funded by the Korea government(MSIT)(RS-2024-00358935). (Corresponding author: Jinsun Park.)}
\thanks{Ukcheol Shin is with the Robotics Institute, School of Computer Science, Carnegie Mellon University, Pittsburgh, Pennsylvania, United States of America (e-mail: ushin@andrew.cmu.edu).} 
\thanks{Jinsun Park is with the School of Computer Science and Engineering, Pusan National University, Busan, Republic of Korea (e-mail: jspark@pusan.ac.kr).} 
\thanks{Color versions of one or more figures in this article are available at
https://doi.org/xx.xxxx/TIV.xxxx.xxxxxxx.}
\thanks{Digital Object Identifier xx.xxxx/TIV.xxxx.xxxxxxx}
}
        
\markboth{Transaction of XXX, ~Vol.~XX, No.~X, September~2024}%
{Deep Depth Estimation from Thermal Image: Dataset, Benchmark, and Challenges}

\maketitle

\begin{abstract}
Achieving robust and accurate spatial perception under adverse weather and lighting conditions is crucial for the high-level autonomy of self-driving vehicles and robots. 
However, existing perception algorithms relying on the visible spectrum are highly affected by weather and lighting conditions.
A long-wave infrared camera (\ie, thermal imaging camera) can be a potential solution to achieve high-level robustness. 
However, the absence of large-scale datasets and standardized benchmarks remains a significant bottleneck to progress in active research for robust visual perception from thermal images.
To this end, this manuscript provides a large-scale Multi-Spectral Stereo (MS$^2$) dataset that consists of stereo RGB, stereo NIR, stereo thermal, stereo LiDAR data, and GNSS/IMU information along with semi-dense depth ground truth.
MS$^2$ dataset includes 162K synchronized multi-modal data pairs captured across diverse locations (\eg, urban city, residential area, campus, and high-way road) at different times (\eg, morning, daytime, and nighttime) and under various weather conditions (\eg, clear-sky, cloudy, and rainy).
Secondly, we conduct a thorough evaluation of monocular and stereo depth estimation networks across RGB, NIR, and thermal modalities to establish standardized benchmark results on MS$^2$ depth test sets (\eg, day, night, and rainy).
Lastly, we provide in-depth analyses and discuss the challenges revealed by the benchmark results, such as the performance variability for each modality under adverse conditions, domain shift between different sensor modalities, and potential research direction for thermal perception. 
Our dataset and source code are publicly available at \url{https://sites.google.com/view/multi-spectral-stereo-dataset} and \url{https://github.com/UkcheolShin/SupDepth4Thermal}.
\end{abstract}


\section{Introduction}
\label{sec:intro}
\begin{figure*}[t]
\begin{center}
{
\begin{tabular}{c@{\hskip 0.005\linewidth}c}
\multicolumn{2}{c}{\includegraphics[width=0.98\linewidth]{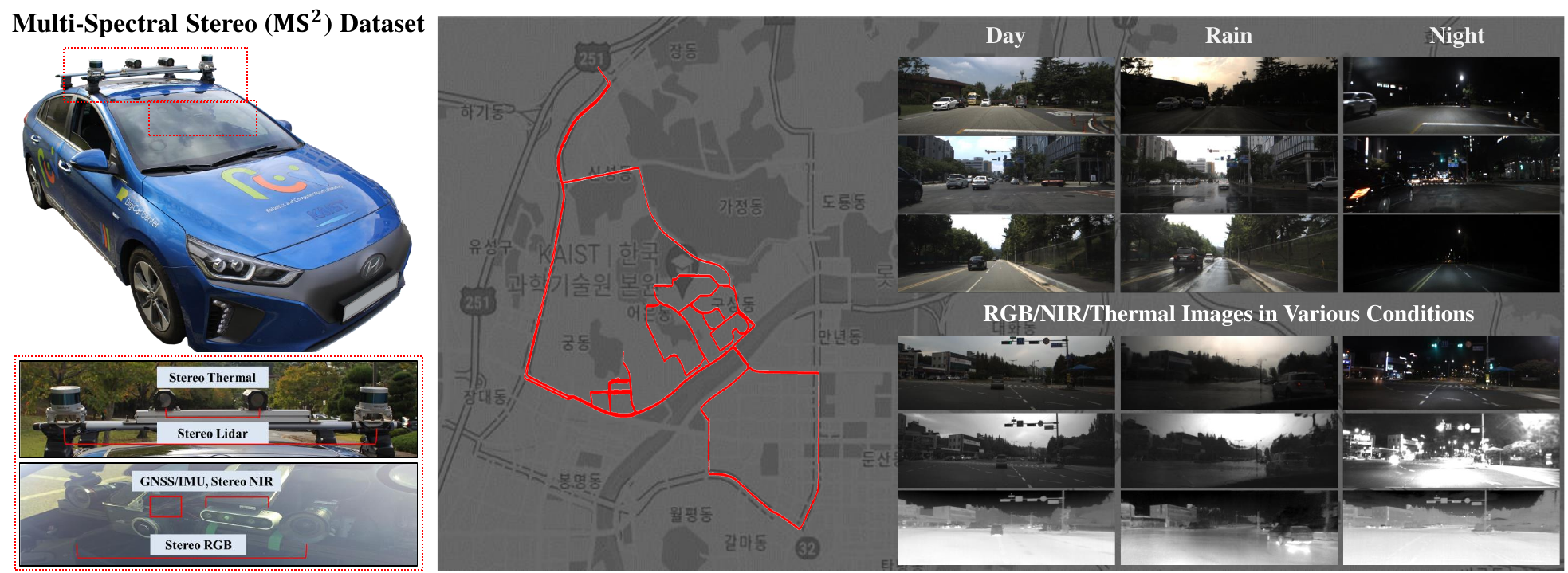}}  \\ 
\multicolumn{2}{c}{{\footnotesize (a) Multi-Spectral Stereo (MS$^2$) dataset overview}} \\ 
\includegraphics[width=0.49\linewidth]{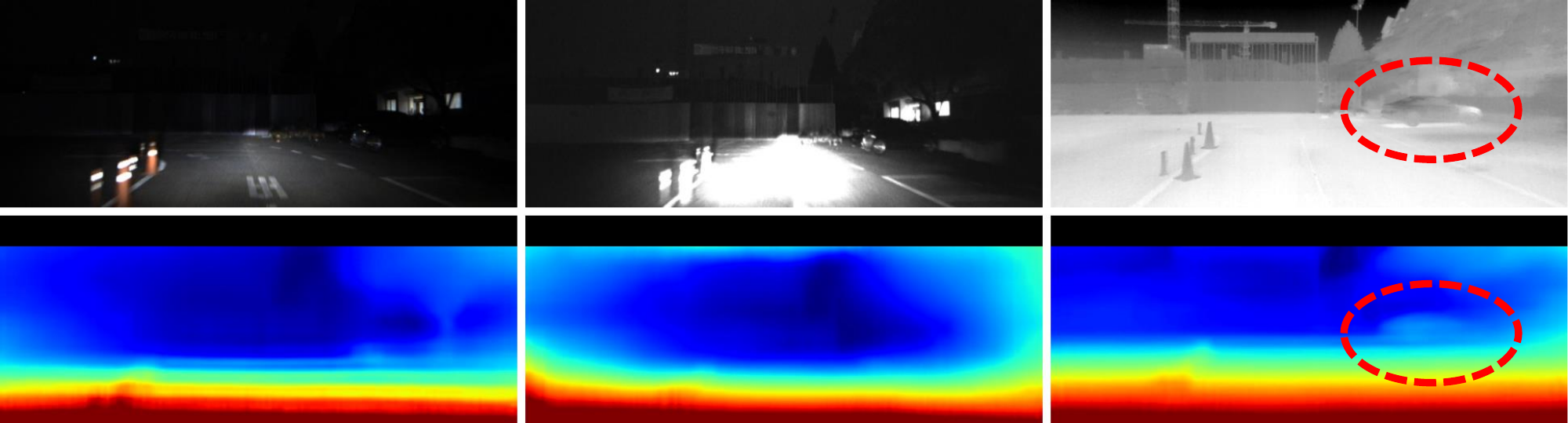} &  
\includegraphics[width=0.49\linewidth]{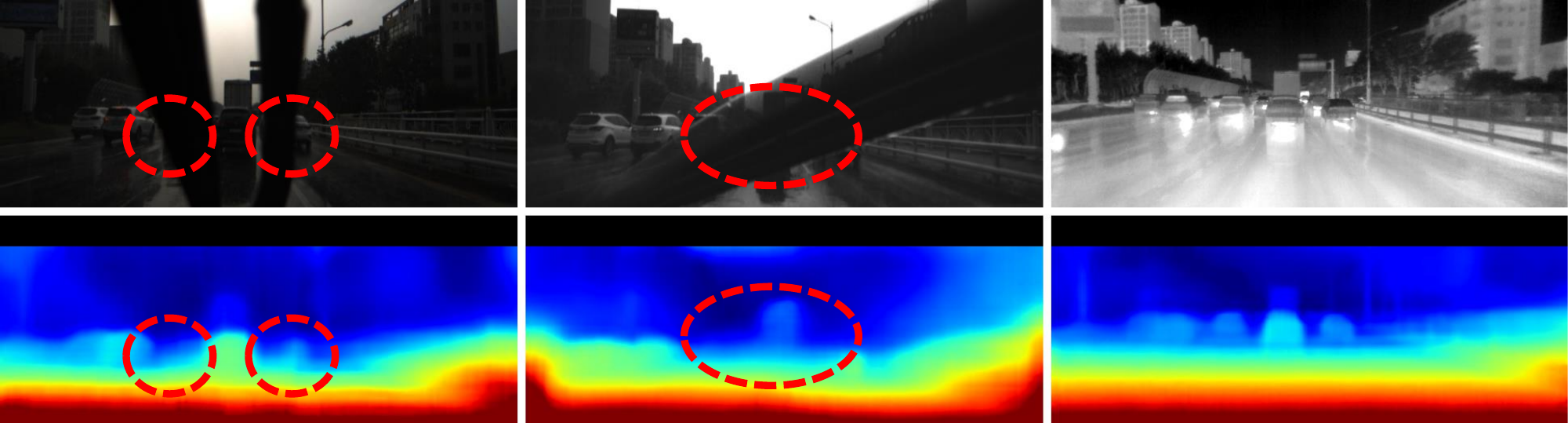} \\
\multicolumn{2}{c}{{\footnotesize (b) Depth from RGB, NIR, and thermal images (Left: night, Right: rainy scenes)}} \\
\end{tabular}
}
\end{center}
\vspace{-0.1in}
\caption{{\bf Overview of Multi-Spectral Stereo (MS$^2$) dataset and depth maps from RGB, NIR, and thermal images in low-visibility conditions.} 
MS$^2$ dataset provides multi-modal stereo data stream, including stereo RGB, stereo NIR, stereo thermal, stereo LiDAR data, and GNSS/IMU information along with semi-dense depth ground truth, captured across diverse locations (\eg, urban city, residential area, campus, and high-way road) at different times (\eg, morning, daytime, and nighttime) and under various weather conditions (\eg, clear-sky, cloudy, and rainy).
Furthermore, depth estimation results from thermal images show high-level reliability and robustness under low-light and rainy conditions.
}
\label{fig:teaser}
\vspace{-0.1in}
\end{figure*}

AUTONOMOUS driving aims to develop intelligent vehicles capable of perceiving their surrounding environments, understanding current contextual information, and making decisions to drive safely without human intervention.
Recent advancements in autonomous vehicles, such as Tesla and Waymo, have been driven by deep neural networks and large-scale vehicular datasets, such as KITTI~\cite{geiger2013vision}, DDAD~\cite{packnet}, and nuScenes~\cite{caesar2020nuscenes}.
These datasets have played a crucial role in developing various perception algorithms, such as depth~\cite{ranftl2020towards}, disparity~\cite{xu2022attention}, object detection~\cite{brazil2023omni3d}, and segmentation~\cite{xiao2024segment} by providing standardized evaluation frameworks for the research community.  
However, a major drawback of existing vehicular datasets is their reliance on visible-spectrum images, which are easily affected by weather and lighting conditions such as rain, fog, dust, haze, and low light. 

Therefore, recent research has actively explored alternative sensors such as Near-Infrared (NIR) cameras~\cite{park2022adaptive}, LiDARs~\cite{guizilini2021sparse,tang2020learning}, radars~\cite{long2021radar,gasperini2021r4dyn}, and long-wave infrared (LWIR) cameras~\cite{lu2021alternative,shin2021self} to achieve reliable and robust visual perception in adverse weather and lighting conditions.
Among these sensors, LWIR camera (\ie, thermal camera) has gained popularity because of its competitive price, adverse weather robustness, and unique modality information (\ie, temperature).
Thermal cameras can capture emissive thermal radiation by objects instead of reflected visible light, making them robust to lighting conditions (\eg, low-light and light flickering) and atmospheric obstructions (\eg, smoke, fog, and rain). 
Therefore, numerous perception algorithms leveraging thermal images~\cite{kim2021ms,sun2020fuseseg,shin2021self,khattak2020keyframe,shin2019sparse} have attracted significant attention to achieve high-level robustness.

However, despite these promising advances, the one missing cornerstone for further active academic research is the well-established large-scale dataset.
Most publicly available datasets for autonomous driving predominantly rely on visible-spectrum (RGB) images, rather than incorporating additional spectral bands such as NIR and LWIR bands.
Especially, despite the advantage of the LWIR band, just a few LWIR datasets have been recently released. 
However, these datasets are indoor oriented~\cite{alee-2019-icra-ws,li2022odombeyondvision,dai2021multi}, small scale~\cite{treible2017cats, alee-2019-icra-ws}, publicly unavailable~\cite{choi2018kaist}, or limited sensor diversity~\cite{lee2022vivid++,choi2018kaist}. 
Therefore, there is a growing need for a large-scale and multi-sensor driving dataset that can support various research to explore the feasibility and challenges associated with a visual perception using multi-spectral sensors.

Another missing cornerstone is the lack of thorough validation for vision perception algorithms in the LWIR spectrum, especially compared to RGB modality under various conditions.
Depth estimation from monocular and stereo images is a fundamental task to perceive and reconstruct the surrounding 3D environments, yet most recent studies in this area have mainly focused on RGB images. 
There are lots of unsolved questions, such as whether the estimated depth map from thermal images is truly better than RGB's in low-visibility conditions and what the challenges are in thermal images.
For instance, compared to RGB images, thermal images typically have lower resolution, less texture, and more noise. 
So, it could pose challenges for a stereo-matching problem despite its robustness against lighting and weather conditions.
These factors might impact the performance of stereo-matching algorithms, making it uncertain whether existing methods will effectively work in the thermal domain. 
As a result, rigorous benchmarking and thorough analysis are essential to assess, improve, and further explore the reliability and robustness of thermal vision perception.

In this manuscript, we provide 1) a large-scale and multi-spectral stereo dataset, named MS$^2$ dataset, along with semi-dense depth Ground Truth (GT) as shown in \cref{fig:teaser} and \cref{sec:dataset}, 2) exhaustive benchmarking results of monocular and stereo depth estimation networks for RGB, NIR, and thermal images on MS$^2$ dataset in \cref{sec:experiments}, and 3) in-depth analysis and challenges that share our findings and introduce numerous research topics in \cref{sec:discussion} to achieve high-level performance, reliability, and robustness against hostile conditions.

Compared to our conference paper~\cite{shin2023deep}, this manuscript newly provides 1) in-depth dataset details, such as sequence information, trajectory visualization, and dataset generation process, 2) additional filtered GT depth and disparity maps for RGB, NIR, and thermal images, 3) exhaustive comparison of depth estimation performance between RGB, NIR, and thermal images, and 4) in-depth analyses and challenges to encourage visual perception research using multi-sensor data.

%
Our contributions can be summarized as follows:
\begin{itemize}
\item 
We provide a large-scale Multi-Spectral Stereo (MS$^2$) dataset, including stereo RGB, stereo NIR, stereo thermal, stereo LiDAR data, and GNSS/IMU data along with ground-truth depth maps.
Our dataset provides about 162K synchronized multi-modal data pairs collected across diverse locations (\eg, urban city, residential area, campus, and high-way road) at different times (\eg, morning, daytime, and nighttime) and under various weather conditions (\eg, clear-sky, cloudy, and rainy).
\item 
We conduct a thorough evaluation of monocular and stereo depth estimation networks across RGB, NIR, and thermal modalities to establish standardized benchmark results on MS$^2$ depth test sets (\eg, day, night, and rainy).
\item
Lastly, we provide in-depth analyses and key challenges that share our findings and introduce numerous research topics, such as the performance degradation for each sensor under adverse conditions, domain shift across different sensor modalities, and potential research direction for thermal perception.
\end{itemize}

\section{Related Work} 
\label{sec:related works}
\subsection{Thermal Image Dataset for 3D Vision}

A well-established large-scale dataset is the most fundamental and top priority for modern deep neural network training.
For the visible spectrum band, numerous large-scale datasets paired with ground-truth depth, disparity, and odometry have been proposed, such as KITTI~\cite{geiger2013vision}, DDAD~\cite{packnet},  Oxford~\cite{maddern20171}, and nuScenes~\cite{caesar2020nuscenes} datasets.
On the other hand, the InfraRed (IR) spectrum band (\eg, near-IR, short-wave IR, long-wave IR) is very rarely included in just a few datasets in a limited form despite its superior environmental robustness. 

The comprehensive comparison is shown in ~\cref{tab:dataset_comparison}.
CATS~\cite{treible2017cats} dataset is a pioneering dataset that provides dense depth GT maps for both indoor and outdoor scenes.
KAIST~\cite{choi2018kaist} and ViViD++ \cite{lee2022vivid++} datasets provide RAW thermal sensory data captured with a vehicle sensor platform. 
OdomBeyondVision \cite{li2022odombeyondvision} and SubT-MRS \cite{zhao2023subtmrs} provide diverse platform data such as a handheld, unmanned ground vehicle (UGV), legged robot, and unmanned aerial vehicle (UAV) in the wild indoor and outdoor scenarios.
However, the above datasets only include single thermal camera views. So, it limits the diverse 3D vision researchers and the nature of stereo vision. 
STheReO \cite{yun2022sthereo} and FIReStereo~\cite{dhrafani2025firestereo} are recent datasets that provide outdoor stereo thermal images in urban cities and wild outdoor environments. 

However, most datasets are still insufficient to investigate the 3D vision research from multi-spectrum image sensors under diverse outdoor driving scenarios.
More specifically, these datasets are indoor oriented~\cite{alee-2019-icra-ws,li2022odombeyondvision,dai2021multi}, small scale~\cite{treible2017cats, alee-2019-icra-ws}, publicly unavailable~\cite{choi2018kaist}, limited in sensor diversity~\cite{lee2022vivid++,choi2018kaist}, limited in weather condition~\cite{alee-2019-icra-ws,choi2018kaist,lee2022vivid++}, or lack of pre-processed dense depth GT~\cite{lee2022vivid++,li2022odombeyondvision,yun2022sthereo,zhao2023subtmrs}.
Therefore, we built a multi-spectral stereo sensor dataset that includes diverse sensor modalities, weather conditions, stereo cues, and GT labels for odometry, dense depth, and disparity maps to investigate monocular and stereo 3D perception research from multi-modal sensors. 


\begin{table*}[t]
\centering
\caption{\textbf{Comprehensive comparison of multi-modal datasets}. 
}
\vspace{-0.10in}
\begin{center}
\resizebox{0.99\linewidth}{!}
{
    \def\arraystretch{1.3}
    \footnotesize
    \begin{tabular}{|c|c|c|c|c|c|c|c|c|c|c|c|c||c|c|c|c|c|} 
    \toprule
    \multirow{2}{*}{Dataset} & \multirow{2}{*}{Year} & \multirow{2}{*}{Environment}  & \multirow{2}{*}{Platform} &  \multirow{2}{*}{LiDAR} & \multirow{2}{*}{IMU} & \multicolumn{2}{c|}{RGB} & \multicolumn{2}{c|}{NIR} & \multicolumn{3}{c|}{Thermal} & \multicolumn{3}{c|}{Weather} & \multicolumn{2}{c|}{Grond-truth} \\ 
    \cline{7-18}
    & & & &  & & Mono & Stereo & Mono & Stereo & Mono & Stereo & RAW & Day & Night & Rain & Odom & Depth \\
    \hline \hline
    CATS~\cite{treible2017cats} & 2017 & In/Outdoor & Handheld  & \cmark & \cmark & \cmark & \cmark & \xmark & \xmark & \cmark & \cmark & \cmark & \cmark & \cmark & \xmark & \xmark & \cmark \\ \hline 
    KAIST~\cite{choi2018kaist} & 2018 & Outdoor & Vehicle  & \cmark & \cmark & \cmark & \cmark & \xmark & \xmark & \cmark & \xmark & \cmark & \cmark & \cmark & \xmark & \cmark & \xmark \\ \hline
    ViViD~\cite{alee-2019-icra-ws} &  2019 & In/Outdoor & Handheld   & \cmark & \cmark & \cmark & \xmark & \xmark & \xmark & \cmark & \xmark & \cmark & \cmark & \cmark & \xmark & \cmark & \xmark \\ \hline
    MultiSpectralMotion~\cite{dai2021multi} & 2021 & In/Outdoor & Handheld & \cmark & \cmark & \cmark & \xmark & \cmark & \xmark & \cmark & \xmark & \cmark & \cmark & \cmark & \xmark & \cmark & \xmark \\ \hline
    ViViD++~\cite{lee2022vivid++} &  2022 & Outdoor & Vehicle & \cmark & \cmark & \cmark & \xmark & \xmark & \xmark & \cmark & \xmark & \cmark & \cmark & \cmark & \xmark & \cmark & \xmark \\ \hline
    \multirow{2}{*}{OdomBeyondVision~\cite{li2022odombeyondvision}} & \multirow{2}{*}{2022} & \multirow{2}{*}{Indoor} & Handheld/  & \multirow{2}{*}{\cmark} & \multirow{2}{*}{\cmark} & \multirow{2}{*}{\cmark} & \multirow{2}{*}{\cmark} & \multirow{2}{*}{\cmark} & \multirow{2}{*}{\cmark} & \multirow{2}{*}{\cmark} & \multirow{2}{*}{\xmark} & \multirow{2}{*}{\cmark} & \multirow{2}{*}{\cmark} & \multirow{2}{*}{\cmark} & \multirow{2}{*}{\xmark} & \multirow{2}{*}{\cmark} & \multirow{2}{*}{\xmark} \\ 
     &  &  & UGV/UAV  &    &  &  &  &  &  &  &  &  &  &  &  &  &  \\ \hline
    STheReO~\cite{yun2022sthereo} & 2022 & Outdoor & Vehicle & \cmark & \cmark & \cmark & \cmark & \xmark & \xmark & \cmark & \cmark & \cmark & \cmark & \cmark & \xmark & \cmark & \xmark \\\hline
    \multirow{2}{*}{SubT-MRS~\cite{zhao2023subtmrs}} & \multirow{2}{*}{2024} & \multirow{2}{*}{In/Outdoor}  & Legged robot/  & \multirow{2}{*}{\cmark} & \multirow{2}{*}{\cmark} & \multirow{2}{*}{\xmark} & \multirow{2}{*}{\cmark} & \multirow{2}{*}{\xmark} & \multirow{2}{*}{\xmark} & \multirow{2}{*}{\cmark} & \multirow{2}{*}{\xmark} & \multirow{2}{*}{\cmark} & \multirow{2}{*}{\cmark} & \multirow{2}{*}{\cmark} & \multirow{2}{*}{\cmark} & \multirow{2}{*}{\cmark} & \multirow{2}{*}{\xmark} \\
     &  &  & UGV/UAV  &    &  &  &  &  &  &  &  &  &  &  &  &  &  \\ \hline
    FIReStereo~\cite{dhrafani2025firestereo} & 2025 & Outdoor & UAV & \cmark &  \cmark &  \xmark &  \xmark &  \xmark &  \xmark & \cmark &  \cmark &  \cmark & \cmark & \cmark & \cmark & \cmark & \cmark \\ 
    \hline
    Ours & 2023 & Outdoor & Vehicle & \cmark &  \cmark &  \cmark &  \cmark &  \cmark &  \cmark & \cmark &  \cmark &  \cmark & \cmark & \cmark & \cmark & \cmark & \cmark \\
    \bottomrule
    \end{tabular}
}
\vspace{-0.2in}
\end{center}
\label{tab:dataset_comparison}
\end{table*}

\subsection{Depth From Visible Spectrum Band}

{\bf Monocular Depth Estimation (MDE)} has high-level universality because it estimates depth map from a single image.
There have been numerous mainstream methods formulating depth estimation as per-pixel regression~\cite{lee2019big,ranftl2020towards, ranftl2021vision, yuan2022neural} by directly estimating per-pixel depth value through a neural network, per-pixel classification~\cite{diaz2019soft, fu2018deep} by discretizing continuous depth range into discrete intervals, and classification-and-regression problems~\cite{bhat2021adabins, li2022binsformer}. 

However, MDE is an ill-posed problem; a single 2D image can be generated from an infinite number of distinct 3D scenes.
Therefore, the estimated monocular depth map is inherently scale-ambiguous, has low generalization performance, and provides lower performance than depth estimation from stereo and multi-view images. 

{\bf Stereo Depth Estimation (SDE)} can estimate metric-scale depth map by utilizing a known camera baseline and disparity map from a rectified stereo image pair.
Existing stereo matching networks can be categorized into 3D cost volume~\cite{liang2018learning, tonioni2019real, xu2020aanet} and 4D cost volume based methods~\cite{chang2018pyramid,guo2019group,shen2021cfnet,xu2022attention}.
The former one estimates a single-channel cost volume (\eg, D$\times$H$\times$W) by measuring the similarity between left and right features. Then, they aggregate the contextual information via 2D convolution.
These methods have high memory and computational efficiency, yet the encoded volume loses large content information, leading to unsatisfactory accuracy. 

The latter one builts multiple-channel cost volume (\eg, D$\times$C$\times$H$\times$W) by concatenating two left-right feature volumes~\cite{chang2018pyramid}, correlation-volume and left-right features~\cite{guo2019group}, or attention-added features~\cite{xu2022attention}. Then, they aggregate the 4D cost volume with 3D convolution layers. 
Current state-of-the-art models are mostly based on this method.
However, this demands high memory consumption and cubic computational complexity that is expensive to deploy in a real-world application.
The SDE task yields significant performance gains compared to the MDE task, yet the SDE task is still struggling to find accurate corresponding points in inherently ill-posed regions such as occlusion areas, repeated patterns, textureless regions, and reflective surfaces.

\subsection{Depth From Thermal Spectrum Band}
Thermal spectrum band has high-level robustness against various adverse weather and lighting conditions, such as rain, fog, dust, haze, and low-light conditions.
However, due to the absence of a large-scale dataset, most previous studies on spatial understanding~\cite{khattak2020keyframe,shin2019sparse,delaune2019thermal,nagase2022shape} have been conducted on their own testbed. 
Additionally, most works focus on utilizing a thermal camera along with other heterogeneous sensors for the target geometric task rather than focusing on the thermal camera itself.

For the spatial understanding task that utilizes a deep neural network, a few researches~\cite{kim2018multispectral,lu2021alternative,shin2021self,shin2022maximize,shin2023self} have been proposed recently.
Most studies focus on the self-supervised depth estimation from thermal images with auxiliary modality guidance, such as aligned-and-paired RGB images~\cite{kim2018multispectral}, style transfer network~\cite{lu2021alternative}, paired RGB images~\cite{shin2021self}, and unpaired RGB image~\cite{shin2023self}.
Unlike the previous studies, in this paper, we target a supervised depth estimation from a single and stereo thermal image that has not yet been actively explored.
 
\begin{figure*}[t]
\centering
\begin{tabular}{c}
\includegraphics[width=0.98\linewidth]{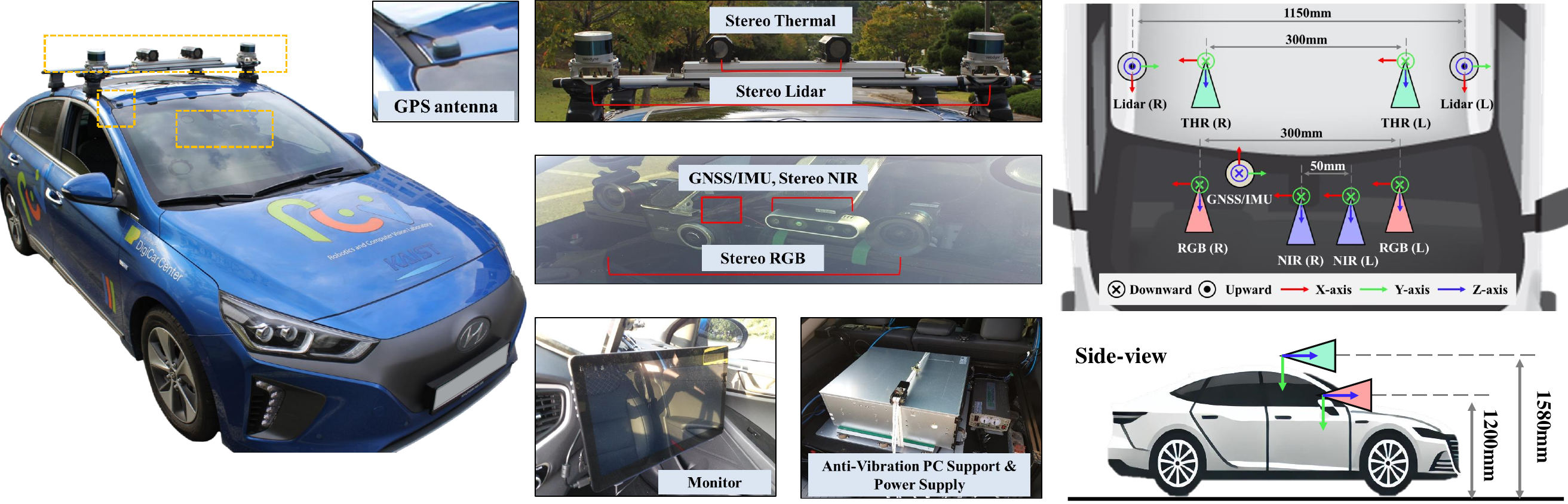} \\
\end{tabular}
\caption{{\bf Overview of vehicular data collection platform for MS$^2$ dataset.} 
We designed a data collection platform consisting of RGB, NIR, thermal, and LiDAR stereo systems and a GPS/IMU module. Stereo RGB, NIR, and IMU modules are installed inside the vehicle to ensure reliable operation under adverse weather conditions. Stereo thermal cameras and LiDARs covered with water-proof housing are built on the vehicle's rooftop.
}
\label{fig:sensor_overview}
\vspace{-0.1in}
\end{figure*}

\section{Multi-Spectral Stereo (MS$^2$) Dataset} 
\label{sec:dataset}
\subsection{Multi-Spectral Stereo Sensor System}

Despite the advantages of the long-wave infrared camera (\ie, thermal camera)~\cite{zhang2021autonomous}, the absence of a large-scale dataset still interrupts the development and investigation of condition-agnostic autonomous driving perception systems from the thermal spectrum domain.
To this end, we designed a data collection platform that consists of RGB, NIR, thermal, and LiDAR stereo system along with a GNSS/IMU module. 
Each sensor specification is described in~\cref{tab:sensor_spec}.

\textbf{System Configuration.}
As shown in~\cref{fig:sensor_overview}, RGB stereo, NIR stereo, and GNSS/IMU module are installed inside the vehicle to ensure safe and reliable operation under adverse weather conditions such as rain, snow, fog, and haze.
Since thermal cameras cannot see through glass, LiDAR stereo, thermal stereo, and GPS antenna are mounted on the vehicle rooftop.
LiDARs are water-proof, and thermal cameras are enclosed in a water-proof housing.
Captured data is transmitted to a local computer via USB 3.0 and Ethernet interfaces. 
The system includes a dedicated power supply with an extra battery and a power inverter. 
Additionally, data collection can be monitored and managed through a monitor panel.

\textbf{Synchronization.}
Accurate time-synchronization is one crucial prerequisite for various 3D geometry tasks with multiple sensors, such as depth estimation, odometry, 3D detection, and 3D reconstruction.
Therefore, we synchronize RGB and NIR stereo cameras using an external synchronizer module that generates a signal at 15 fps. 
Thermal stereo cameras are also synchronized at 15 fps with a sync signal generated by the left thermal camera.
Additionally, a software trigger ensures synchronization between the two systems at the start of each data acquisition.
As a result, RGB, NIR, and thermal stereo data are captured at 15 fps with synchronized signals. 
Meanwhile, stereo LiDARs are synchronized using the Pulse Per Second (PPS) signal from the GNSS/IMU module to acquire synchronized point clouds and minimize interference between LiDARs during the scanning process.

\begin{table}[t]
\centering
\caption{\textbf{Sensor specification for data collection platform}. 
} \vspace{-0.1in}
\begin{center}
\resizebox{0.99\linewidth}{!}
{
    \def\arraystretch{1.2}
    \footnotesize
    \begin{tabular}{c|ccc} 
    \toprule
    Sensor & Model & Frame Rate & Characteristics \\
    \hline
    \multirow{3}{*}{RGB camera} & PointGrey BlackFly-S & \multirow{2}{*}{Max 75 fps} &  2448$\times$2048 pixel \\ 
    & BFS-U3-51S5C & & Global Shutter \\ \cline{2-4}
    & Kowa LM5JC10M & & 82.2$^\circ$ (H) $\times$ 66.5$^\circ$(V) FoV \\ \hline
    \multirow{3}{*}{NIR camera} & \multirow{3}{*}{Intel RealSense D435i} & \multirow{3}{*}{Max 90 fps} & 1280 $\times$ 800 pixel\\ 
    & & & Global Shutter \\ 
    & & & 91$^\circ$ (H) $\times$ 65$^\circ$(V) FoV \\ \hline
    \multirow{4}{*}{Thermal camera} & \multirow{4}{*}{FLIR A65C} & \multirow{4}{*}{Max 30 fps} & 640$\times$512 pixel \\
    & & & 90$^\circ$ (H) $\times$ 69$^\circ$(V) FoV   \\
    & & & Uncooled VOX microbolometer  \\
    & & & 16-bit Raw data  \\ \hline
    \multirow{3}{*}{LiDAR} & \multirow{3}{*}{Velodyne VLP-16} & \multirow{3}{*}{Max 20 fps} & Accuracy: $\pm$ 3 cm  \\
    & & & Measurement range : 100m \\ 
    & & & 360$^\circ$ (H), $\pm$15$^\circ$ (V) FoV \\ \hline
    \multirow{2}{*}{GNSS/IMU} & LORD Microstrain & \multirow{2}{*}{10/100 Hz} & Position, Velocity, \\
                              & 3DM-GX5-45       & & Attitude, Acceleration, etc.\\
    \bottomrule
    \end{tabular} 
}
\vspace{-0.2in}
\end{center}
\label{tab:sensor_spec}
\end{table}

\begin{figure*}[t!]
\begin{center}
{
\begin{tabular}{c@{\hskip 0.005\linewidth}c}
\includegraphics[width=0.49\linewidth]{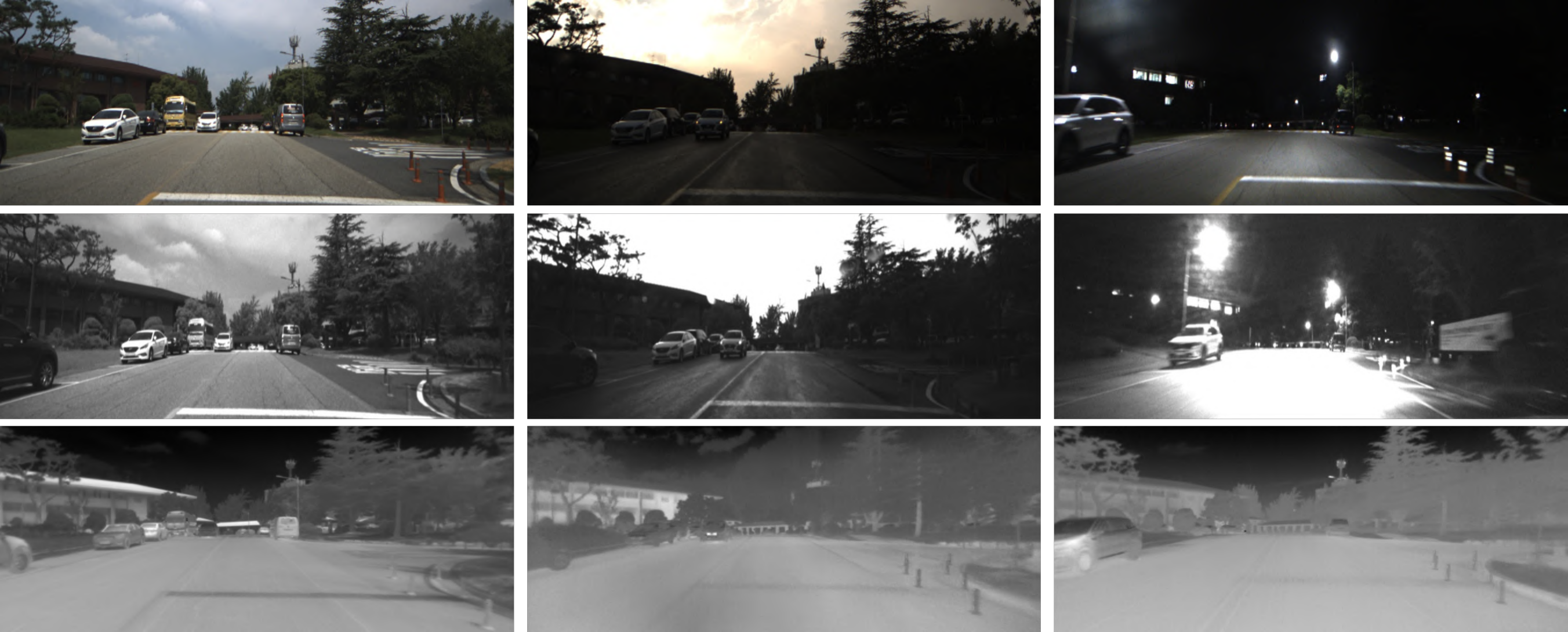} &
\includegraphics[width=0.49\linewidth]{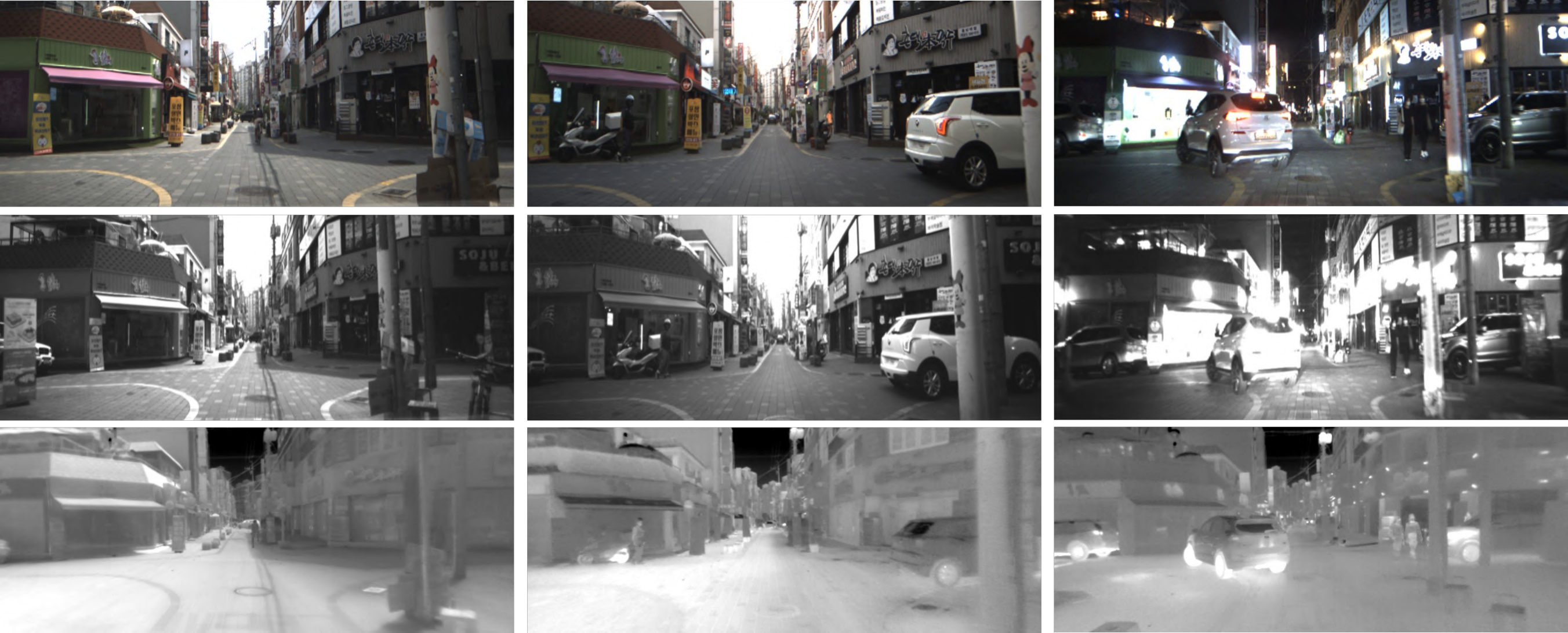} \\
{\footnotesize (a) Driving Scenario - Campus (Day/Cloudy/Night)} & {\footnotesize (b) Driving Scenario - Residential (Day/Cloudy/Night)} \\
\includegraphics[width=0.49\linewidth]{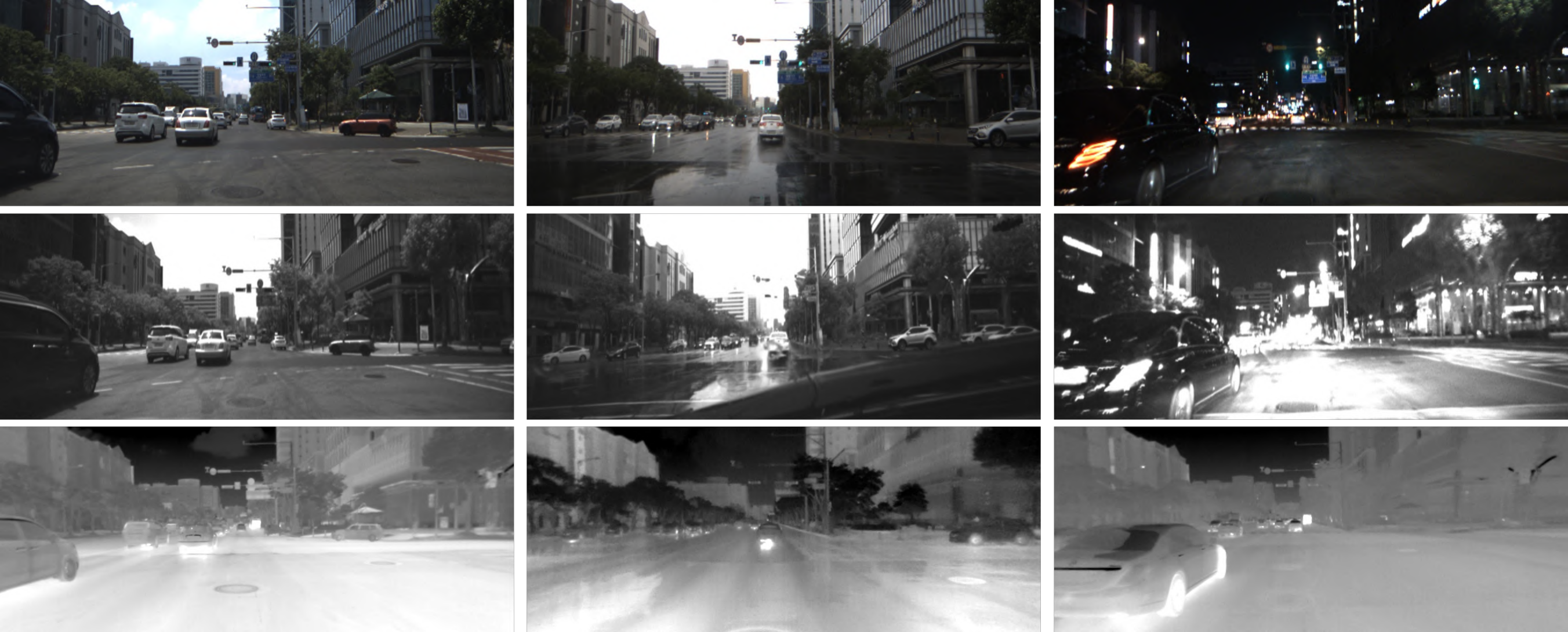} &
\includegraphics[width=0.49\linewidth]{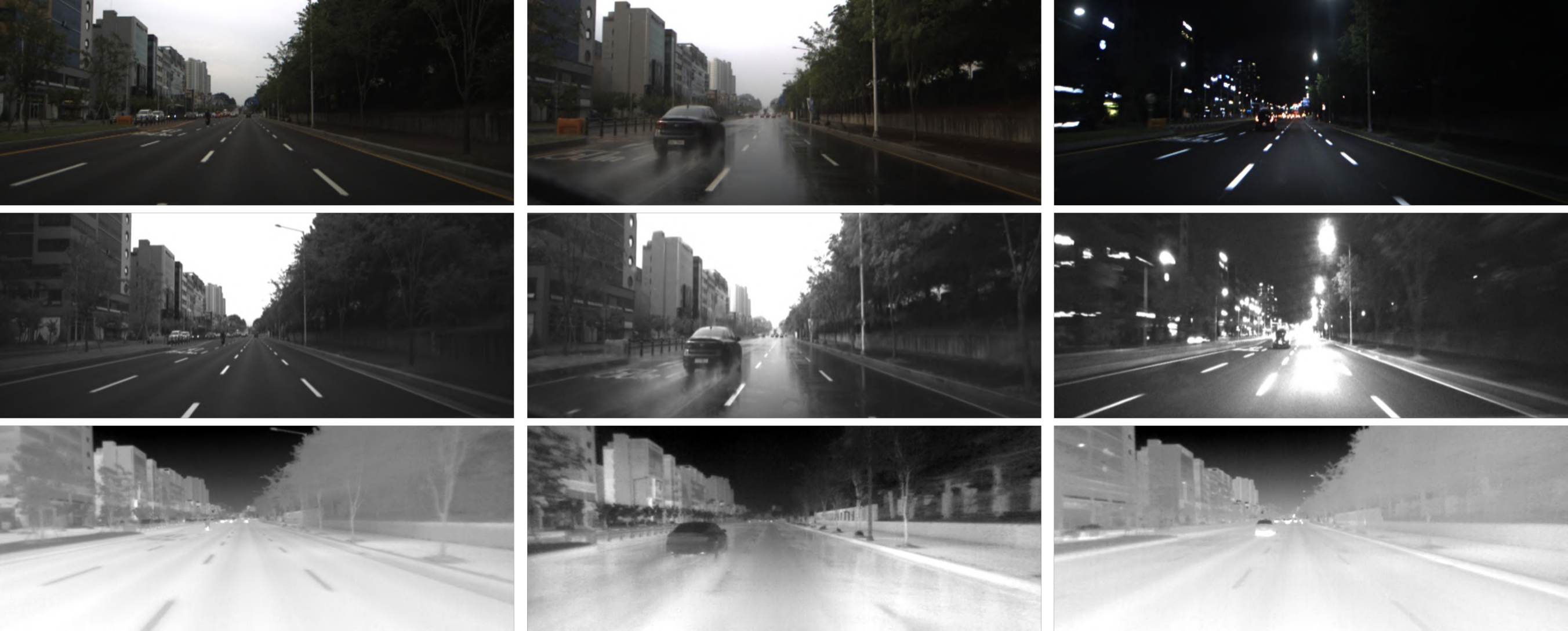} \\
{\footnotesize (c) Driving Scenario - City (Day/Rain/Night)}& {\footnotesize (d) Driving Scenario - Road2,4 (Day/Rain/Night)} \\
\end{tabular}
\begin{tabular}{c@{\hskip 0.005\linewidth}c@{\hskip 0.005\linewidth}c@{\hskip 0.005\linewidth}c}
\includegraphics[width=0.24\linewidth]{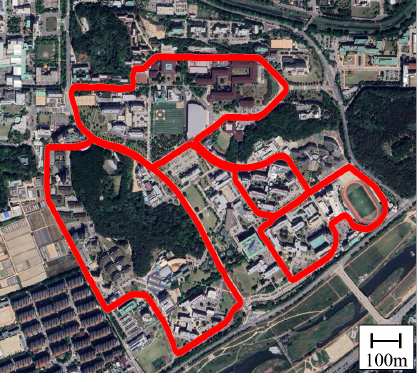} &
\includegraphics[width=0.24\linewidth]{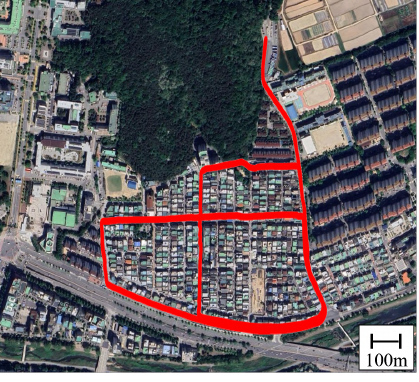} &
\includegraphics[width=0.24\linewidth]{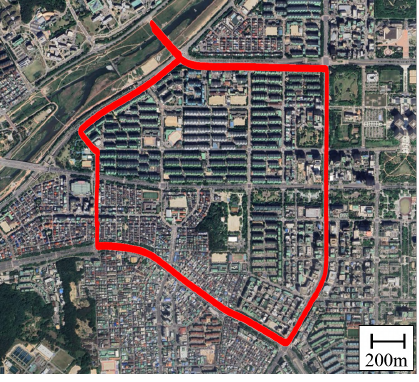} &
\includegraphics[width=0.24\linewidth]{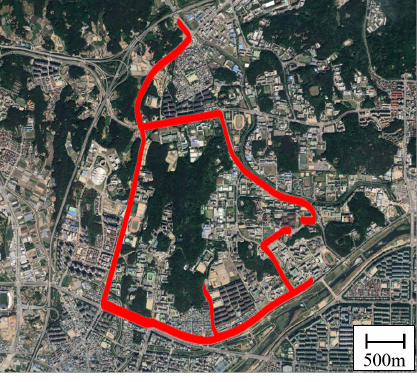} \\
{\footnotesize (e) Campus } & {\footnotesize (f) Residential area} & {\footnotesize (g) Urban area} & {\footnotesize (h) Road-1,2,3,4} \\
\end{tabular}
}
\end{center}
\vspace{-0.1in}
\caption{{\bf Data examples of Multi-Spectral Stereo (MS$^2$) outdoor driving dataset.} 
The collected dataset provides about 162K synchronized data taken under locations of campus, city, residential area, road, and suburban with various time slots (morning, day, and night) and weather conditions (clear-sky, cloudy, and rainy)).
For each block, three rows indicate RGB, NIR, and thermal images, respectively. 
According to the surrounding conditions, each spectrum sensor shows different aspects, advantages, and disadvantages induced by their sensor characteristics).
}
\label{fig:MS2_datset_example}
\vspace{-0.3in}
\end{figure*}

\begin{table*}[bh!]
\caption{\textbf{Sequence list of MS$^2$ dataset.} 
The MS$^2$ dataset provides 20 RAW ROSbag files taken under various locations, times, and weather conditions.
Each ROSbag contains raw RGB stereo, NIR stereo, thermal stereo, LiDAR stereo, and GNSS/IMU data stream.
We can observe diverse characteristics according to the combination of time, weather, and location.
}
\vspace{-0.1in}
\begin{center}
\resizebox{0.95\linewidth}{!}
{
    \def\arraystretch{1.15}
    \footnotesize
    \begin{tabular}{|c|c|c|c|c|c|c|c|c|c|}
    \hline
    Index & Sequence Name & Time & Weather & Loc. & Duration & \# of pairs & Inten. Mean & Temp. Mean & Temp. Std\\ 
    \hline
    \hline
    1   & 2021-08-06-10-59-33 & Morning  & Clear sky     & Campus      & 1071.0 sec & 10,443 & 76.5 & \SI{43.7}{\celsius} & $\pm$ 6.9 \\
    2	& 2021-08-06-17-44-55 & Daytime  & Cloudy\&Rainy & Campus      & 1100.9 sec & 10,764 & 63.6 & \SI{33.3}{\celsius} & $\pm$ 2.1 \\
    3 	& 2021-08-13-17-06-04 & Daytime  & Clear sky     & Campus      & 984.0 sec  & 9,690  & 71.4 & \SI{38.1}{\celsius} & $\pm$ 3.6 \\
    4	& 2021-08-13-21-18-04 & Nightime & Clear sky     & Campus      & 1040.0 sec & 10,132 & 26.8 & \SI{33.3}{\celsius} & $\pm$ 2.3 \\ \hline
    5	& 2021-08-06-11-37-46 & Morning  & Clear sky     & Urban	   & 1118.9 sec & 9,516  & 80.1 & \SI{44.6}{\celsius} & $\pm$ 6.0 \\
    6	& 2021-08-06-16-19-00 & Daytime  & Cloudy\&Rainy & Urban       & 1218.8 sec & 11,869 & 72.7 & \SI{35.4}{\celsius} & $\pm$ 2.4\\
    7	& 2021-08-13-15-46-56 & Daytime  & Clear sky     & Urban       & 1201.0 sec & 11,813 & 68.9 & \SI{39.3}{\celsius} & $\pm$ 4.4 \\
    8	& 2021-08-13-21-36-10 & Nightime & Clear sky     & Urban       & 1212.3 sec & 11,941 & 51.1 & \SI{34.1}{\celsius} & $\pm$ 2.5 \\ \hline
    9	& 2021-08-06-11-23-45 & Morning  & Clear sky     & Residential & 599.9 sec  & 5,811  & 83.8 & \SI{44.5}{\celsius} & $\pm$ 5.3\\
    10	& 2021-08-06-16-45-28 & Daytime  & Cloudy\&Rainy & Residential & 665.5 sec  & 6,673  & 77.2 & \SI{37.4}{\celsius} & $\pm$ 2.8 \\
    11	& 2021-08-13-16-14-48 & Daytime  & Clear sky     & Residential & 929.6 sec  & 9,168  & 76.5 & \SI{39.1}{\celsius} & $\pm$ 3.0 \\
    12	& 2021-08-13-22-03-03 & Nightime & Clear sky     & Residential & 773.9 sec  & 7,552  & 52.5 & \SI{34.6}{\celsius} & $\pm$ 2.2 \\ \hline
    13	& 2021-08-06-16-59-13 & Daytime  & Cloudy\&Rainy & Road1       & 614.2 sec  & 6,484  & 54.9 & \SI{35.7}{\celsius} & $\pm$ 2.4\\
    14	& 2021-08-13-16-31-10 & Daytime  & Clear sky     & Road1       & 579.65 sec & 5,696  & 58.9 & \SI{38.9}{\celsius} & $\pm$ 4.5 \\
    15	& 2021-08-13-22-16-02 & Nightime & Clear sky     & Road1       & 543.47 sec & 5,235  & 35.0 & \SI{33.6}{\celsius} & $\pm$ 2.9 \\ \hline
    16	& 2021-08-06-17-21-04 & Daytime  & Cloudy\&Rainy & Road2       & 1177.9 sec & 11,458 & 63.5 & \SI{33.2}{\celsius} & $\pm$ 1.9\\
    17  & 2021-08-13-16-50-57 & Daytime  & Clear sky     & Road2       & 883.8 sec  & 8,530  & 67.3 & \SI{38.3}{\celsius} & $\pm$ 4.2 \\  \hline
    18	& 2021-08-13-16-08-46 & Daytime  & Clear sky     & Road3       & 259.9 sec  & 2,544  & 65.1 & \SI{39.0}{\celsius} & $\pm$ 4.4 \\
    19	& 2021-08-13-21-58-13 & Nightime & Clear sky     & Road3       & 261.3 sec  & 2,540  & 38.6 & \SI{34.1}{\celsius} & $\pm$ 2.3 \\  \hline
    20	& 2021-08-13-22-36-41 & Nightime & Clear sky     & Road4       & 434.6 sec  & 4,213  & 28.0 & \SI{33.4}{\celsius} & $\pm$ 2.3 \\
    \hline
    \end{tabular}
}
\end{center}
\label{tab:MS_sequence}
\end{table*}

\subsection{Data Collection}
Multi-Spectral Stereo (MS$^2$) dataset collects 20 RAW ROSbag files that are taken under various locations (\eg, urban area, campus, residential area, and road), times (\eg, morning, daytime, and nighttime), and weather conditions (\eg, clear-sky, cloudy, and rainy), as shown in~\cref{fig:MS2_datset_example} and listed in~\cref{tab:MS_sequence}.
Each ROSbag contains raw data streams from RGB stereo, NIR stereo, thermal stereo, LiDAR stereo, and a GNSS/IMU module. 
As shown in~\cref{fig:MS2_datset_example}, the dataset provides diverse temporal and weather variations for the same locations, allowing researchers to analyze the advantages, robustness, and domain gap issue according to environmental condition change for each modality.
For instance, we can easily identify the unique characteristics of each modality, such as light sensitivity in NIR images and light-agnostic nature in thermal images.
Each sequence exhibits varying mean intensity and temperature values depending on its location, weather, and time conditions, ranging from 20 to 80 in intensity and \SI{30}{\celsius} to \SI{45}{\celsius} in temperature.
Additionally, the dataset includes odometry trajectories and GPS information acquired by the GNSS/IMU module, featuring a variety of open loops and single and multiple closed loops. 
Therefore, we believe the dataset can also be used for place recognition and long-term SLAM research across sensors, time, and weather changes.

\subsection{Location and Sensor Characteristics}
Each sequence in~\cref{tab:MS_sequence} has distinct characteristics influenced by time, location, and weather conditions.
Here, we outline the unique properties of each location and the behavior of each sensor under varying lighting and weather conditions.

{\bf Locations.}
\textit{Campus} offers a relatively static driving scenario with a moderate number of parked vehicles, bicycles, motorcycles, and a few moving pedestrians (\cref{fig:MS2_datset_example}-(a)).
At nighttime, it has limited lighting sources, such as streetlights and illuminated buildings, resulting in lower brightness levels (Mean intensity: 26.8) compared to other scenarios.
In contrast, \textit{Urban city} presents a highly dynamic and complex traffic situation with lots of moving objects, including a high density of vehicles and pedestrians, as shown in \cref{fig:MS2_datset_example}-(c).

\textit{Residential area} features a mix of static and dynamic elements, where pedestrians and vehicles often appear suddenly due to occlusion caused by parked vehicles and closely spaced buildings (\cref{fig:MS2_datset_example}-(b)).
Both \textit{Urban city} and \textit{Residential area} have higher brightness levels at nighttime, as they benefit from abundant external light sources, including streetlights, vehicle headlights, and illuminated buildings.
Lastly, \textit{Road} scenario is characterized by high-speed vehicles with relatively few buildings. 
Most objects, such as vehicles, buildings, and streetlights, are sparsely distributed and positioned far from the ego-vehicle. 
As a result, the brightness levels at nighttime are relatively lower than \textit{Urban city} and \textit{Residential area}.

{\bf Lighting Condition.}
As shown in~\cref{fig:MS2_datset_example}, each sensor captures different aspects of the environment depending on lighting conditions.
RGB images can provide detailed object textures, color, and sharp structure information under well-lit conditions.
However, as the lighting condition gets worse (\ie, nighttime), this information is often limited, saturated, or blurred. 
On the other hand, NIR images provide relatively better image quality in low-light conditions (\eg, nighttime in \textit{Road} scenarios) because the NIR spectrum is more sensitive to the light source.
However, this property also leads to frequent saturation in car head-light and street lamp regions.
Thermal images remain unaffected by lighting conditions, as thermal imaging is based on heat signatures rather than visible light. 
This lighting-agnostic property makes thermal sensors particularly useful in environments with extreme lighting variations.

{\bf Weather Condition.}
Weather conditions affect RGB and NIR images in similar ways. Under clear daytime sky, both exhibit high contrast and sharp details, resulting in clear image quality. 
However, in rainy conditions, water droplets on the lens cause light scattering and blurring. 
Additionally, windshield wiper movements introduce intermittent occlusions, as shown in~\cref{fig:teaser}-(b).

While thermal images are unaffected by lighting conditions, they are influenced by weather conditions, particularly in terms of heat sources. 
Under the clear daytime sky, thermal images display high contrast and clean edge information because the sun acts as an external heat source, providing strong thermal radiation to all objects. 
Depending on their heat absorption and reflectance properties, objects emit distinct thermal energy, enabling the thermal camera to capture high-contrast images (\eg, daytime thermal images in~\cref{fig:MS2_datset_example}-(a-d)).
However, in the absence of a strong heat source (\ie, nighttime) or under heavy cloud cover (\ie, cloudy/rainy conditions), thermal image contrast decreases as the thermal radiation differences between objects become less pronounced.
%
%

\begin{figure}[t]
\begin{center}
{
\begin{tabular}{c@{\hskip 0.005\linewidth}c@{\hskip 0.005\linewidth}c}
\includegraphics[width=0.28\linewidth]{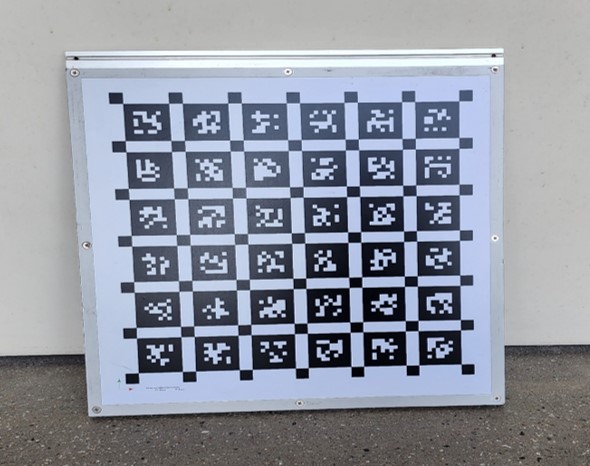} & 
\includegraphics[width=0.28\linewidth]{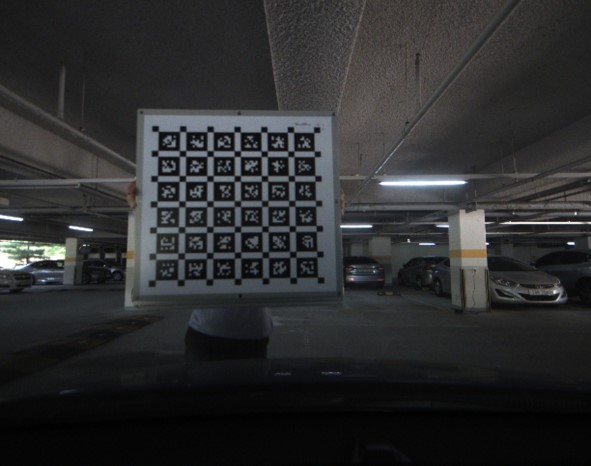} & 
\includegraphics[width=0.28\linewidth]{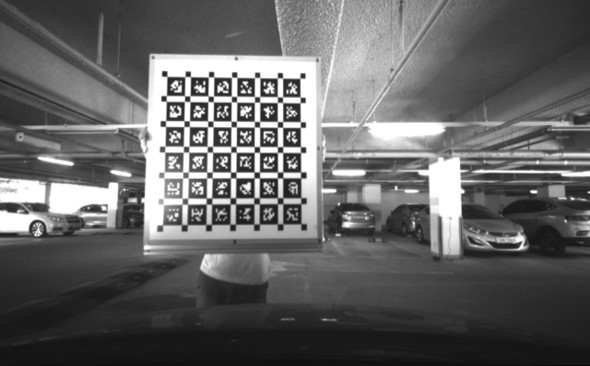} \\  
{\footnotesize (a) AprilTag (6x6)} & {\footnotesize (b) RGB image} & {\footnotesize (c) NIR image} \\
\includegraphics[width=0.28\linewidth]{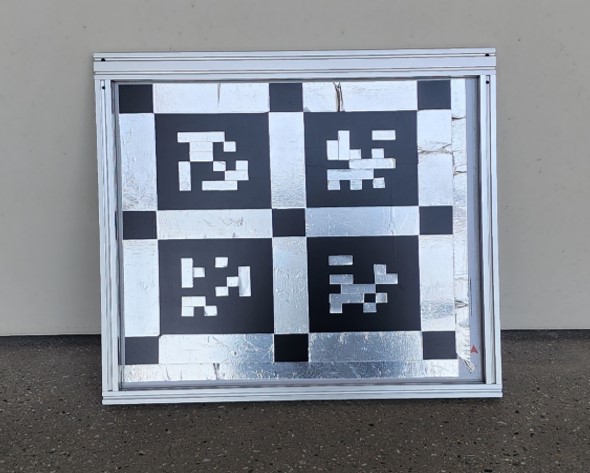} & 
\includegraphics[width=0.28\linewidth]{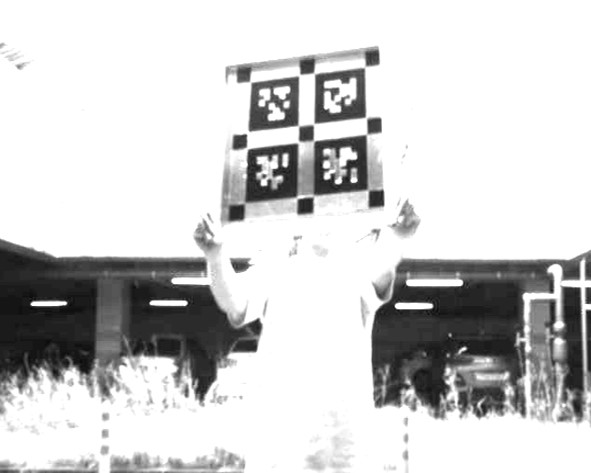} & 
\includegraphics[width=0.28\linewidth]{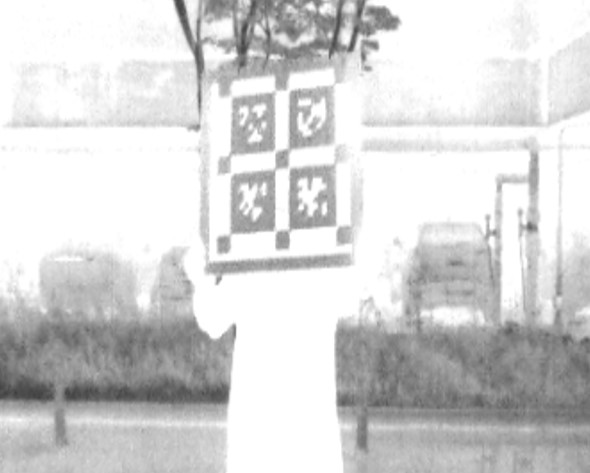} \\
{\footnotesize (d) AprilTag (2x2)} & {\footnotesize (e) NIR image} & {\footnotesize (f) THR image} \\
\includegraphics[width=0.28\linewidth]{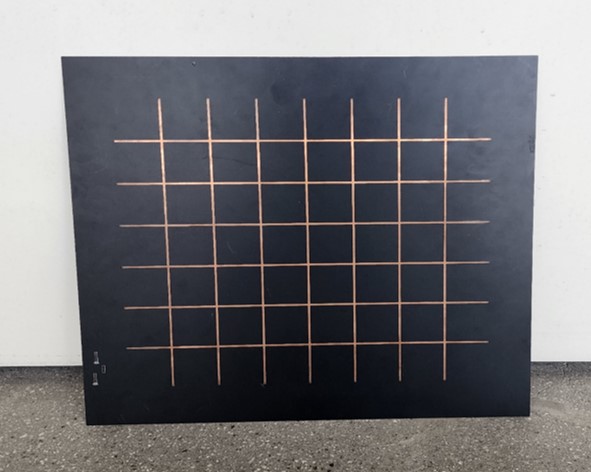} & 
\includegraphics[width=0.28\linewidth]{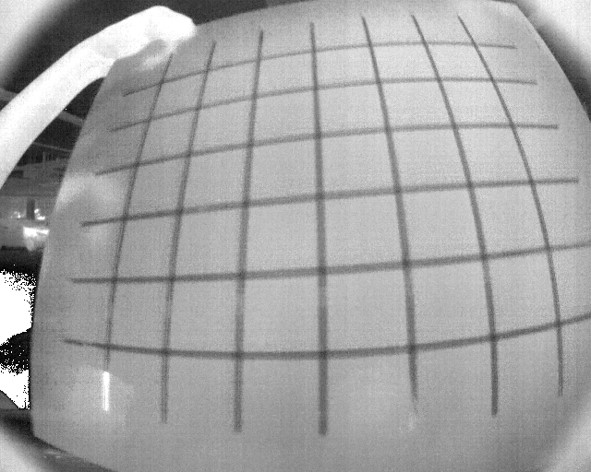} & 
\includegraphics[width=0.28\linewidth]{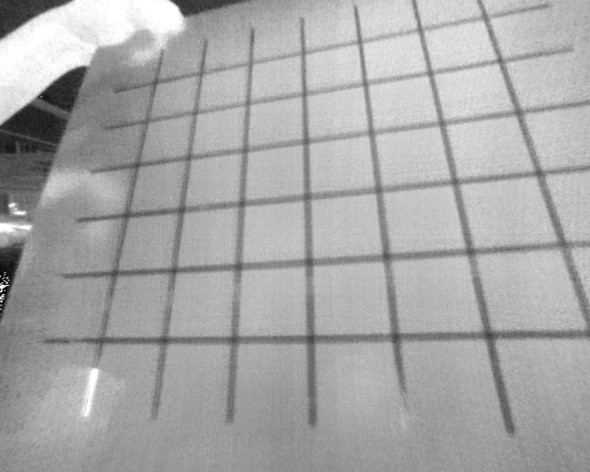} \\
{\footnotesize (g) Line-board } & {\footnotesize (h) THR image} & {\footnotesize (i) Rectified image} \\
\end{tabular}
}
\end{center}
\vspace{-0.1in}
\caption{{\bf Pattern boards for multi-sensor calibration}.
We utilize three pattern boards (\ie, 6x6 AprilTag, 2x2 AprialTag, and copper-coated line boards)  for multi-sensor calibrations.
}
\label{fig:ms2_cal_board}
\vspace{-0.10in}
\end{figure}

\subsection{Calibration}
The MS$^2$ dataset provides intrinsic and extrinsic parameters for all sensors to make our dataset applicable to various computer vision tasks.
To this end, we use a 6×6 AprilTag~\cite{wang2016apriltag} board for stereo RGB, stereo NIR, RGB-NIR, NIR-IMU, and NIR-LiDAR calibrations~\cite{zhang2000flexible,rehder2016extending}, as shown in~\cref{fig:ms2_cal_board}.
For stereo thermal cameras, we employ a copper-coated line board to estimate intrinsic matrices, radial distortion parameters, and the extrinsic matrix. 
Additionally, we use a 2×2 AprilTag board with metallic tape to estimate the extrinsic matrix between NIR and thermal cameras.
Before capturing pattern board images, both the 2×2 AprilTag board and the line board are cooled down to enhance thermal image contrast between metallic and non-metallic regions.
Once the calibration process is complete, the remaining extrinsic matrices are derived using the matrix multiplications.

\subsection{Image Rectification \& Cropping}

\begin{figure}[t!]
\begin{center}
{
\begin{tabular}{c}
\includegraphics[width=0.98\linewidth]{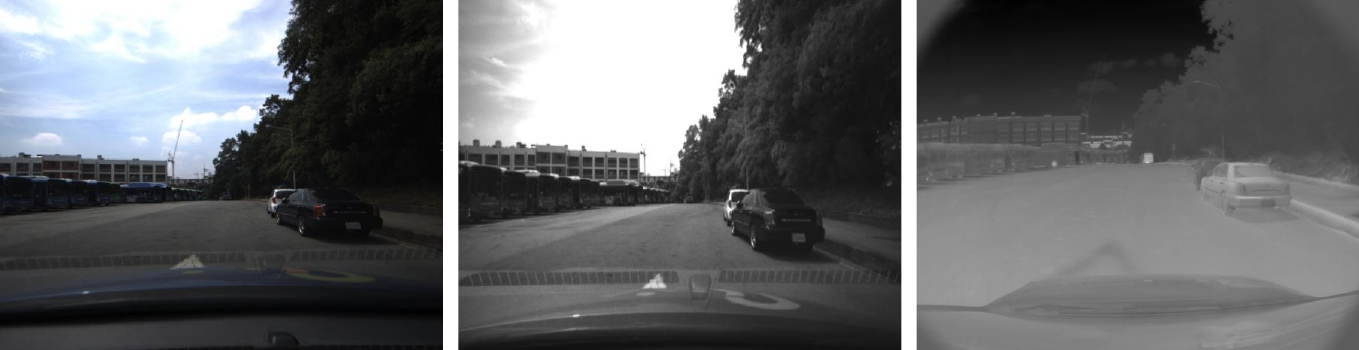} \vspace{-0.05in} \\
{\footnotesize (a) Original Images (RGB/NIR/THR)}\\ 
\includegraphics[width=0.98\linewidth]{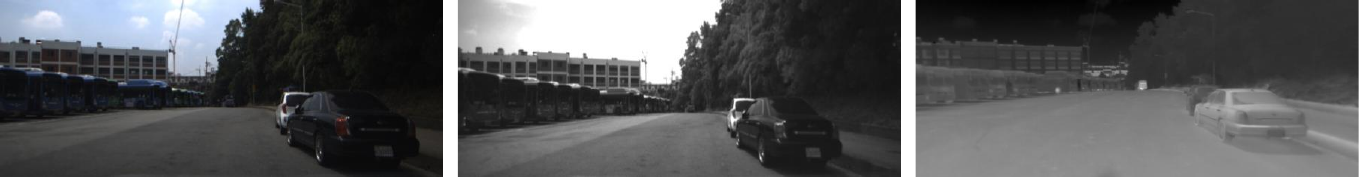} \vspace{-0.05in} \\
{\footnotesize (b) Rectified\&Cropped Images (RGB/NIR/THR)} \\
\end{tabular}
}
\end{center}
\vspace{-0.1in}
\caption{{\bf Image pre-processing for MS$^2$ dataset}.
We rectified and cropped the original RGB, NIR, and thermal images with intrinsic and distortion parameters to make valid training data. 
}
\label{fig:ms2_rectify}
\vspace{-0.1in}
\end{figure}

The original RGB, NIR, and thermal images contain unnecessary regions, such as the car hood and sky, which are irrelevant for many vision applications. 
Additionally, projected LiDAR depth points do not appear in the sky and are invalid on the car hood.
Therefore, we rectified and cropped the RGB, NIR, and thermal images after the calibration process, retaining only the valid regions, as shown in~\cref{fig:ms2_rectify}.
After rectification and cropping, the final spatial resolutions are 1224$\times$384 for RGB, 1280$\times$352 for NIR, and 640$\times$256 for thermal images.

\subsection{Multi-Spectral Stereo (MS$^2$) Depth Dataset}
{\bf MS$^2$-Depth: Ground-Truth Generation Process.}
Dense and accurate GT depth maps are vital to train and test modern deep neural networks. 
Therefore, we delicately generate as dense and accurate GT depth maps as possible by aggregating 32 consecutive stereo LiDAR scans, projecting them onto each image plane, and filtering the projected depth map.
The detailed processes are as follows.

First, we interpolate GNSS/IMU sensor data to estimate the pose of each sensor at every timestamp. 
Then, we accumulate 32 successive stereo LiDAR scans using the extrinsic matrix between left and right LiDARs and transformation matrices between consecutive LiDAR frames. 
After that, we apply the Iterative Closest Point (ICP) algorithm~\cite{besl1992method} to refine the accumulated point clouds and remove overlapping points.
Next, the refined 3D point cloud is projected onto each image plane (\ie, RGB, NIR, and thermal images). 

\begin{figure}[t]
\begin{center}
{
\begin{tabular}{c}
\includegraphics[width=0.95\linewidth]{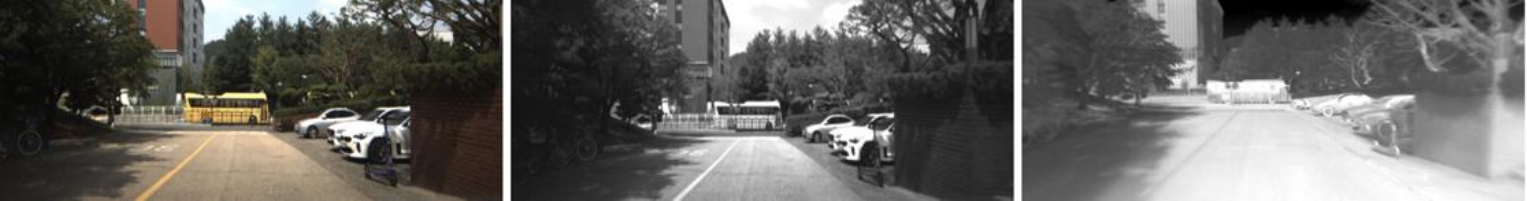} \vspace{-0.05in} \\ 
{\footnotesize (a) Processed images (RGB/NIR/THR)} \\
\includegraphics[width=0.95\linewidth]{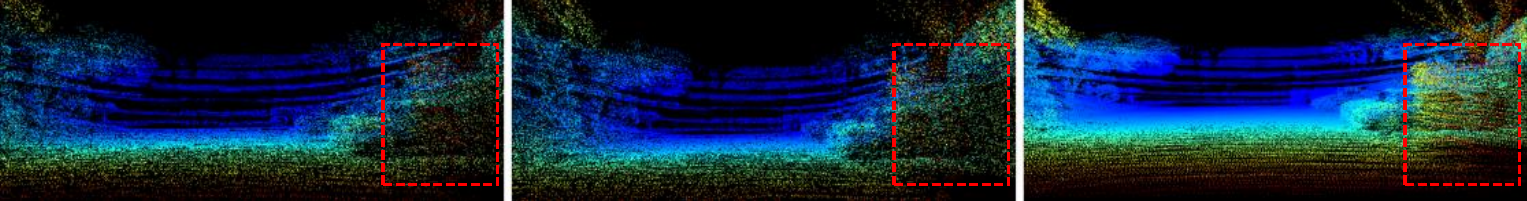} \vspace{-0.05in} \\ 
{\footnotesize (b) Projected LiDAR points} \\
\includegraphics[width=0.95\linewidth]{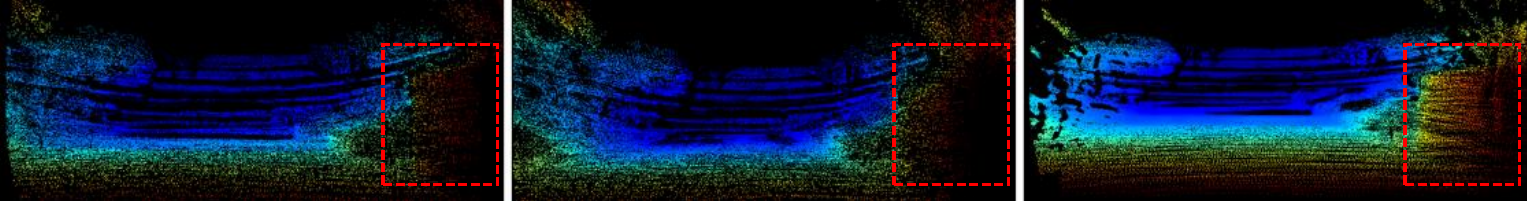} \vspace{-0.05in} \\ 
{\footnotesize (c) Filtered LiDAR points} \\
\end{tabular}
}
\end{center}
\vspace{-0.1in}
\caption{{\bf Depth GT generation process for MS$^2$ depth dataset}.
We merge consecutive 32 frames of stereo LiDAR scans with ICP method~\cite{besl1992method}. After that, we filter out unreliable points with left-right consistency and depth ratio checks. 
}
\label{fig:ms2_filtering}
\vspace{-0.1in}
\end{figure}

However, due to the camera perspective geometry and the sparsity of point clouds, the non-overlapping 3D points appear overlapped in the 2D image plane, as shown in~\cref{fig:ms2_filtering}-(b).
To mitigate this issue, we apply two filtering methods: left-right consistency check~\cite{geiger2013vision} and depth ratio check.
The left-right consistency check reconstructs the left image by warping the right image using a disparity map, which is converted from the projected depth map.
Depth points with significant reconstruction errors (\ie, $|I_{L}-\Tilde{I}_{L}|>Thres$) between the original left ${I}_L$ and synthesized $\Tilde{I}_L$ images are then excluded.

The depth ratio check compares the projected depth map $D_p$ with an estimated depth map $D_e$ using a fine-tuned stereo matching network~\cite{xu2020aanet}\footnote{We fine-tune networks for each modality (RGB, NIR, and thermal images) with carefully selected depth maps.}.
If the depth ratio exceeds a predefined threshold that indicates significant discrepancies (\ie, $max(D_p/D_e,D_e/D_p)=\delta > Thres$), the corresponding depth points are discarded.
After two processes, we finally get the filtered depth maps as shown in~\cref{fig:ms2_filtering}-(c).


{\bf MS$^2$-Depth: Training, Validation, and Evaluation Sets.}
From the MS$^2$ dataset, we periodically sampled RGB, NIR, thermal images, and depth GT pairs to construct training, validation, and evaluation splits for the benchmarks of monocular and stereo depth networks. 
In total, the training set contains 26K pairs, and the validation set has 4K pairs. 
For evaluation, we use 2.3K pairs for daytime, 2.3K for nighttime, and 2.5K for rainy conditions.
We try to make the training, validation, and evaluation splits have almost zero overlaps in time, weather, and location diversity.

Specifically, the training set is composed of sequences from “Road2” (\cref{tab:MS_sequence}, Seq. 16-17), “Road4” (Seq. 20), “Urban” (Seq. 5,8), and “Campus” (Seq. 1-3). 
In contrast, all sequences from “Residential,” “Road1,” and “Road3” are reserved for validation and evaluation.
Additionally, “Urban” (Seq. 6-7) and “Campus” (Seq. 4), which feature different lighting and weather conditions from the training set, are allocated to validation and testing. 
Lastly, considering lighting and weather conditions, the evaluation data is subdivided into three categories: daytime (Seq. 7,9,14), nighttime (Seq. 4,12,15), and rainy (Seq. 6,10,13). 
The remaining sequences (Seq. 11,18,19) are used for validation.

\section{MS$^2$-Depth Benchmark Results}
\label{sec:experiments}

\subsection{Implementation Details}
\label{subsec:impl_detail}

{\bf MDE and SDE Networks.}
To validate various MDE and SDE networks for RGB, NIR, and thermal spectrum, we train and evaluate representative models using the proposed MS$^2$-Depth dataset.
Specifically, we adopt MDE networks based on regression~\cite{lee2019big}, classification~\cite{fu2018deep}, classification-and-regression~\cite{bhat2021adabins}, and modern transformer architectures~\cite{yuan2022neural} (\ie, BTS, DORN, AdaBins, and NeWCRF). 
Also, we employ SDE networks utilizing 3D cost volume~\cite{chang2018pyramid,xu2020aanet} and 4D cost volume~\cite{guo2019group,xu2022attention} (\ie, PSMNet, AANet, GwcNet, and ACVNet).
We utilize their official source code to implement each network architecture. 
All networks are initialized with ImageNet pretrained~\cite{deng2009imagenet} or provided backbone models by following their original implementations. 

{\bf Optimizer and Data Augmentation.}  
All models are trained for 60 epochs on a single A6000 GPU with 48GB of memory.
We use a batch size of 8 for MDE models and 4 for SDE models. 
Training is conducted using the AdamW optimizer with an initial learning rate of $1e^{-4}$. 
Cosine Annealing Warm Restarts is used for learning rate scheduling.
For data augmentation, we apply random center crop-and-resize, brightness jitter, and contrast jitter to all models. 
Horizontal flip is additionally applied to the MDE networks.

\begin{table*}[t!]
\caption{\textbf{Benchmarking results of MDE networks on the MS$^2$ depth dataset across RGB, NIR, and thermal modalities}. 
The table presents benchmarking results of representative MDE networks~\cite{fu2018deep,lee2019big,bhat2021adabins,yuan2022neural} with different input modalities (a) RGB, (b) NIR, and (c) thermal on the MS$^2$ depth evaluation sets (\ie, daytime, nighttime, and rainy).
The best performance in each modality is highlighted in \textbf{bold}. 
}
\begin{center}
\subfloat[Monocular depth estimation results on the MS$^2$ depth evaluation sets (\textbf{\textit{Modality: RGB images}}).]
{ 
    \resizebox{0.90 \linewidth}{!}{
    \def\arraystretch{0.9}
    \footnotesize
    \begin{tabular}{|c|c|cccc|ccc|}
    \hline
    \multirow{2}{*}{Methods} & \multirow{2}{*}{TestSet} & \multicolumn{4}{c|}{\textbf{Error $\downarrow$}} & 
    \multicolumn{3}{c|}{\textbf{Accuracy $\uparrow$}}  \\ \cline{3-9}
     &  &  AbsRel & SqRel & RMSE & RMSElog &  $\delta < 1.25$ & $\delta < 1.25^{2}$ & $\delta < 1.25^{3}$ \\ 
    \hline \hline
    \multirow{4}{*}{DORN~\cite{fu2018deep}} 
    & Day   & 0.107 & 0.639 & 4.206 & 0.149 & 0.897 & 0.982 & 0.995\\
    & Night & 0.129 & 0.750 & 4.547 & 0.170 & 0.830 & 0.974 & 0.995\\
    & Rain  & 0.154 & 1.302 & 5.698 & 0.203 & 0.796 & 0.955 & 0.987\\ \cline{2-9}
    & \cellcolor{Gray1}Avg   & \cellcolor{Gray1}0.130 & \cellcolor{Gray1}0.908 & \cellcolor{Gray1}4.840 & \cellcolor{Gray1}0.175 & \cellcolor{Gray1}0.840 & \cellcolor{Gray1}0.970 & \cellcolor{Gray1}0.992\\ \hline
    \multirow{4}{*}{BTS~\cite{lee2019big}} 
    & Day   & 0.081 & 0.383 & 3.324 & 0.117 & 0.938 & 0.990 & 0.997\\
    & Night & 0.109 & 0.589 & 4.095 & 0.146 & 0.871 & 0.982 & 0.997\\
    & Rain  & 0.130 & 0.911 & 4.908 & 0.174 & 0.843 & 0.969 & 0.991\\ \cline{2-9}
    & \cellcolor{Gray1}Avg   & \cellcolor{Gray1}0.107 & \cellcolor{Gray1}0.635 & \cellcolor{Gray1}4.128 & \cellcolor{Gray1}0.147 & \cellcolor{Gray1}0.883 & \cellcolor{Gray1}0.980 & \cellcolor{Gray1}0.995\\ \hline
    \multirow{4}{*}{AdaBins~\cite{bhat2021adabins}} 
    & Day   & 0.093 & 0.442 & 3.578 & 0.128 & 0.921 & 0.988 & 0.997\\
    & Night & 0.111 & 0.598 & 4.043 & 0.147 & 0.869 & 0.981 & 0.997\\
    & Rain  & 0.131 & 0.891 & 4.915 & 0.174 & 0.843 & 0.968 & 0.991\\ \cline{2-9}
    & \cellcolor{Gray1}Avg   & \cellcolor{Gray1}0.112 & \cellcolor{Gray1}0.650 & \cellcolor{Gray1}4.197 & \cellcolor{Gray1}0.150 & \cellcolor{Gray1}0.877 & \cellcolor{Gray1}0.979 & \cellcolor{Gray1}0.995\\ \hline   
    \multirow{4}{*}{NeWCRF~\cite{yuan2022neural}} 
    & Day   & 0.077 & 0.333 & 3.111 & 0.107 & 0.948 & 0.994 & 0.999\\
    & Night & 0.099 & 0.464 & 3.573 & 0.130 & 0.899 & 0.989 & 0.998\\
    & Rain  & 0.119 & 0.745 & 4.447 & 0.158 & 0.870 & 0.977 & 0.994\\ \cline{2-9}
    & \cellcolor{Gray1}Avg   & \cellcolor{Gray1}\textbf{0.099} & \cellcolor{Gray1}\textbf{0.520} & \cellcolor{Gray1}\textbf{3.729} & \cellcolor{Gray1}\textbf{0.133} & \cellcolor{Gray1}\textbf{0.905} & \cellcolor{Gray1}\textbf{0.987} & \cellcolor{Gray1}\textbf{0.997}\\  \hline
    \end{tabular}
    }
}
\vspace{3mm}
\subfloat[Monocular depth estimation results on the MS$^2$ depth evaluation sets (\textbf{\textit{Modality: NIR images}}).]
{ 
    \resizebox{0.90 \linewidth}{!}{
    \def\arraystretch{0.9}
    \footnotesize
    \begin{tabular}{|c|c|cccc|ccc|}
    \hline
    \multirow{2}{*}{Methods} & \multirow{2}{*}{TestSet} & \multicolumn{4}{c|}{\textbf{Error $\downarrow$}} & 
    \multicolumn{3}{c|}{\textbf{Accuracy $\uparrow$}}  \\ \cline{3-9}
     &  &  AbsRel & SqRel & RMSE & RMSElog &  $\delta < 1.25$ & $\delta < 1.25^{2}$ & $\delta < 1.25^{3}$ \\ 
    \hline \hline
    \multirow{4}{*}{DORN~\cite{fu2018deep}} 
    & Day   & 0.112 & 0.507 & 3.677 & 0.147 & 0.883 & 0.986 & 0.997 \\
    & Night & 0.123 & 0.623 & 3.750 & 0.154 & 0.856 & 0.981 & 0.996\\
    & Rain  & 0.180 & 1.567 & 5.872 & 0.226 & 0.744 & 0.937 & 0.980\\ \cline{2-9}
    & \cellcolor{Gray1}Avg   & \cellcolor{Gray1}0.139 & \cellcolor{Gray1}0.917 & \cellcolor{Gray1}4.471 & \cellcolor{Gray1}0.177 & \cellcolor{Gray1}0.825 & \cellcolor{Gray1}0.967 & \cellcolor{Gray1}0.991\\ \hline
    \multirow{4}{*}{BTS~\cite{lee2019big}} 
    & Day   & 0.094 & 0.408 & 3.396 & 0.126 & 0.917 & 0.991 & 0.998 \\
    & Night & 0.108 & 0.533 & 3.452 & 0.136 & 0.893 & 0.983 & 0.996\\
    & Rain  & 0.164 & 1.365 & 5.518 & 0.209 & 0.783 & 0.944 & 0.982\\ \cline{2-9}
    & \cellcolor{Gray1}Avg   & \cellcolor{Gray1}0.123 & \cellcolor{Gray1}0.784 & \cellcolor{Gray1}4.159 & \cellcolor{Gray1}0.158 & \cellcolor{Gray1}0.862 & \cellcolor{Gray1}0.972 & \cellcolor{Gray1}0.992\\ \hline
    \multirow{4}{*}{AdaBins~\cite{bhat2021adabins}} 
    & Day   & 0.092 & 0.394 & 3.351 & 0.123 & 0.917 & 0.990 & 0.999 \\
    & Night & 0.107 & 0.523 & 3.411 & 0.135 & 0.894 & 0.984 & 0.997\\
    & Rain  & 0.160 & 1.260 & 5.311 & 0.201 & 0.790 & 0.950 & 0.986\\ \cline{2-9}
    & \cellcolor{Gray1}Avg   & \cellcolor{Gray1}0.121 & \cellcolor{Gray1}0.740 & \cellcolor{Gray1}4.059 & \cellcolor{Gray1}0.154 & \cellcolor{Gray1}0.865 & \cellcolor{Gray1}0.974 & \cellcolor{Gray1}0.993\\ \hline
    \multirow{4}{*}{NeWCRF~\cite{yuan2022neural}} 
    & Day   & 0.086 & 0.333 & 3.071 & 0.113 & 0.933 & 0.994 & 0.999 \\
    & Night & 0.099 & 0.448 & 3.157 & 0.124 & 0.912 & 0.987 & 0.998\\
    & Rain  & 0.150 & 1.104 & 5.042 & 0.191 & 0.810 & 0.958 & 0.987\\ \cline{2-9}
    & \cellcolor{Gray1}Avg   & \cellcolor{Gray1}\textbf{0.112} & \cellcolor{Gray1}\textbf{0.641} & \cellcolor{Gray1}\textbf{3.791} & \cellcolor{Gray1}\textbf{0.144} & \cellcolor{Gray1}\textbf{0.883} & \cellcolor{Gray1}\textbf{0.979} & \cellcolor{Gray1}\textbf{0.994}\\  \hline
    \end{tabular}
    }
}
\vspace{3mm}
\subfloat[Monocular depth estimation results on the MS$^2$ depth evaluation sets (\textbf{\textit{Modality: thermal images}}).]
{ 
    \resizebox{0.90 \linewidth}{!}{
    \def\arraystretch{0.9}
    \footnotesize
    \begin{tabular}{|c|c|cccc|ccc|}
    \hline
    \multirow{2}{*}{Methods} & \multirow{2}{*}{TestSet} & \multicolumn{4}{c|}{\textbf{Error $\downarrow$}} & 
    \multicolumn{3}{c|}{\textbf{Accuracy $\uparrow$}} \\ \cline{3-9}
     &  &  AbsRel & SqRel & RMSE & RMSElog &  $\delta < 1.25$ & $\delta < 1.25^{2}$ & $\delta < 1.25^{3}$ \\ 
    \hline \hline
    \multirow{4}{*}{DORN~\cite{fu2018deep}} 
    & Day   & 0.101 & 0.443 & 3.389 & 0.135 & 0.909 & 0.987 & 0.998\\
    & Night & 0.099 & 0.401 & 3.152 & 0.129 & 0.914 & 0.989 & 0.998\\
    & Rain  & 0.127 & 0.757 & 4.377 & 0.167 & 0.842 & 0.971 & 0.994\\ \cline{2-9}
    & \cellcolor{Gray1}Avg   & \cellcolor{Gray1}0.109 & \cellcolor{Gray1}0.540 & \cellcolor{Gray1}3.660 & \cellcolor{Gray1}0.144 & \cellcolor{Gray1}0.887 & \cellcolor{Gray1}0.982 & \cellcolor{Gray1}0.997\\ \hline
    \multirow{4}{*}{BTS~\cite{lee2019big}} 
    & Day   & 0.075 & 0.314 & 2.945 & 0.107 & 0.943 & 0.992 & 0.998\\
    & Night & 0.079 & 0.282 & 2.755 & 0.106 & 0.943 & 0.995 & 0.999\\
    & Rain  & 0.102 & 0.531 & 3.740 & 0.137 & 0.896 & 0.983 & 0.997\\ \cline{2-9}
    & \cellcolor{Gray1}Avg   & \cellcolor{Gray1}0.086 & \cellcolor{Gray1}0.380 & \cellcolor{Gray1}3.163 & \cellcolor{Gray1}0.117 & \cellcolor{Gray1}0.927 & \cellcolor{Gray1}0.990 & \cellcolor{Gray1}0.998\\ \hline
    \multirow{4}{*}{AdaBins~\cite{bhat2021adabins}} 
    & Day   & 0.079 & 0.327 & 3.003 & 0.111 & 0.937 & 0.992 & 0.999\\
    & Night & 0.079 & 0.273 & 2.679 & 0.105 & 0.947 & 0.995 & 0.999\\
    & Rain  & 0.104 & 0.518 & 3.725 & 0.139 & 0.891 & 0.983 & 0.997\\ \cline{2-9}
    & \cellcolor{Gray1}Avg   & \cellcolor{Gray1}0.088 & \cellcolor{Gray1}0.377 & \cellcolor{Gray1}3.152 & \cellcolor{Gray1}0.119 & \cellcolor{Gray1}0.924 & \cellcolor{Gray1}0.990 & \cellcolor{Gray1}0.998\\ \hline
    \multirow{4}{*}{NeWCRF~\cite{yuan2022neural}} 
    & Day   & 0.071 & 0.271 & 2.717 & 0.099 & 0.951 & 0.994 & 0.999\\
    & Night & 0.074 & 0.248 & 2.544 & 0.099 & 0.952 & 0.996 & 1.000\\
    & Rain  & 0.095 & 0.464 & 3.503 & 0.127 & 0.909 & 0.988 & 0.998\\ \cline{2-9}
    & \cellcolor{Gray1}Avg   & \cellcolor{Gray1}\textbf{0.081} & \cellcolor{Gray1}\textbf{0.331} & \cellcolor{Gray1}\textbf{2.937} & \cellcolor{Gray1}\textbf{0.109} & \cellcolor{Gray1}\textbf{0.937} & \cellcolor{Gray1}\textbf{0.992} & \cellcolor{Gray1}\textbf{0.999}\\  \hline
    \end{tabular}
    }
}
\end{center}
\vspace{-0.1in}
\label{table:ms2_bench_mono}
\end{table*}
\begin{figure*}[t]
\begin{center}
{
\begin{tabular}{c@{\hskip 0.01\linewidth}c}
\includegraphics[width=0.49\linewidth]{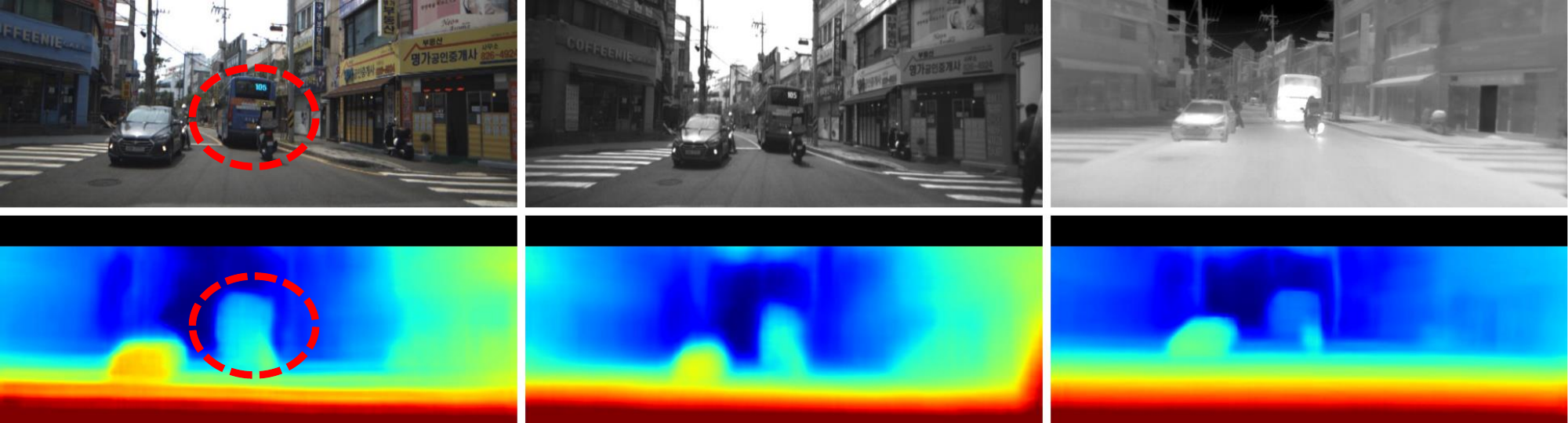} &  \includegraphics[width=0.49\linewidth]{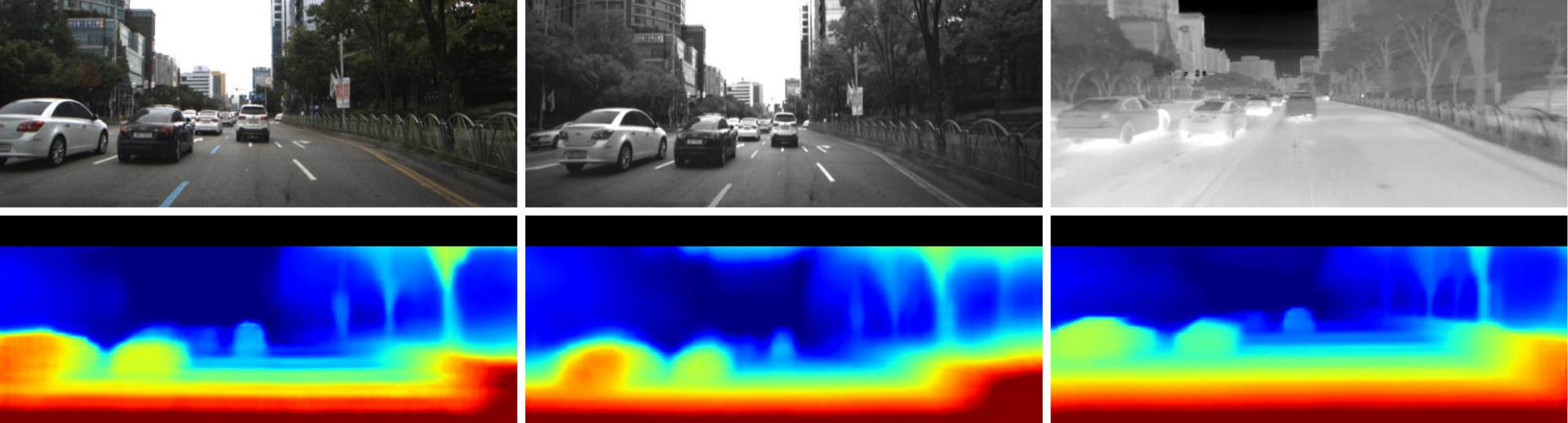} \vspace{-0.02in} \\ 
\multicolumn{2}{c}{{\footnotesize (a) Depth maps from RGB, NIR, and thermal images (MS$^2$ Depth - Daytime)}} \\
\includegraphics[width=0.49\linewidth]{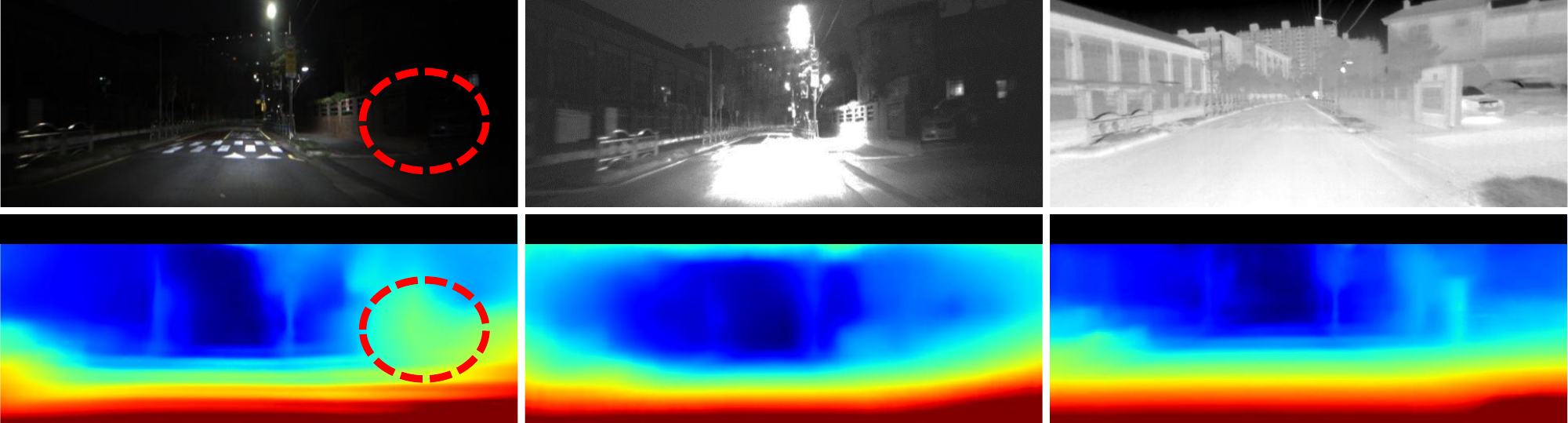} &  \includegraphics[width=0.49\linewidth]{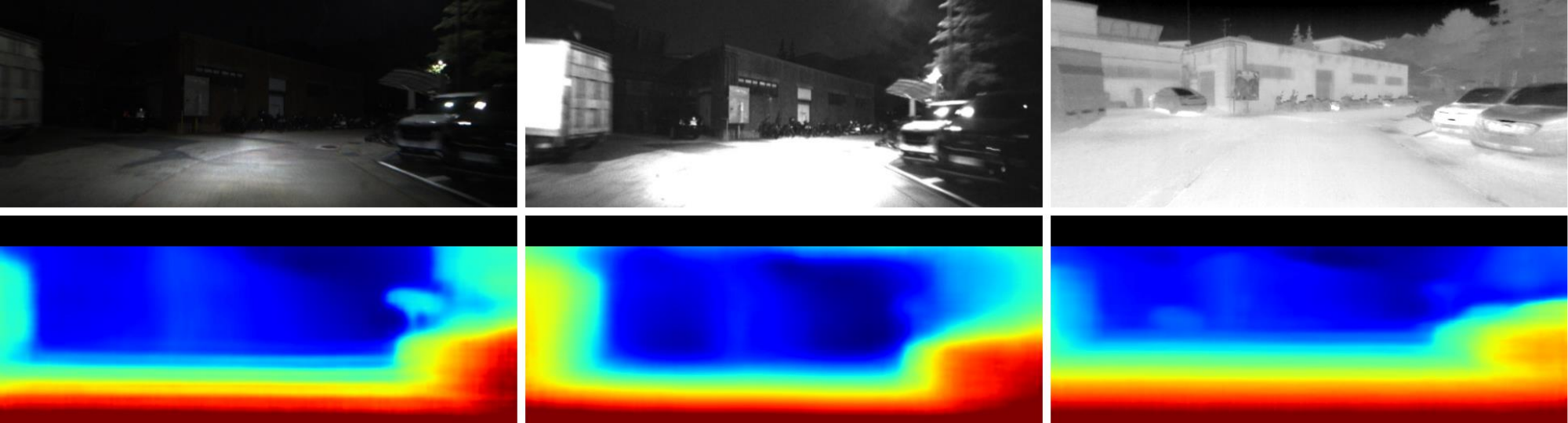} \vspace{-0.02in} \\ 
\multicolumn{2}{c}{{\footnotesize (b) Depth maps from RGB, NIR, and thermal images (MS$^2$ Depth - Nighttime)}} \\
\includegraphics[width=0.49\linewidth]{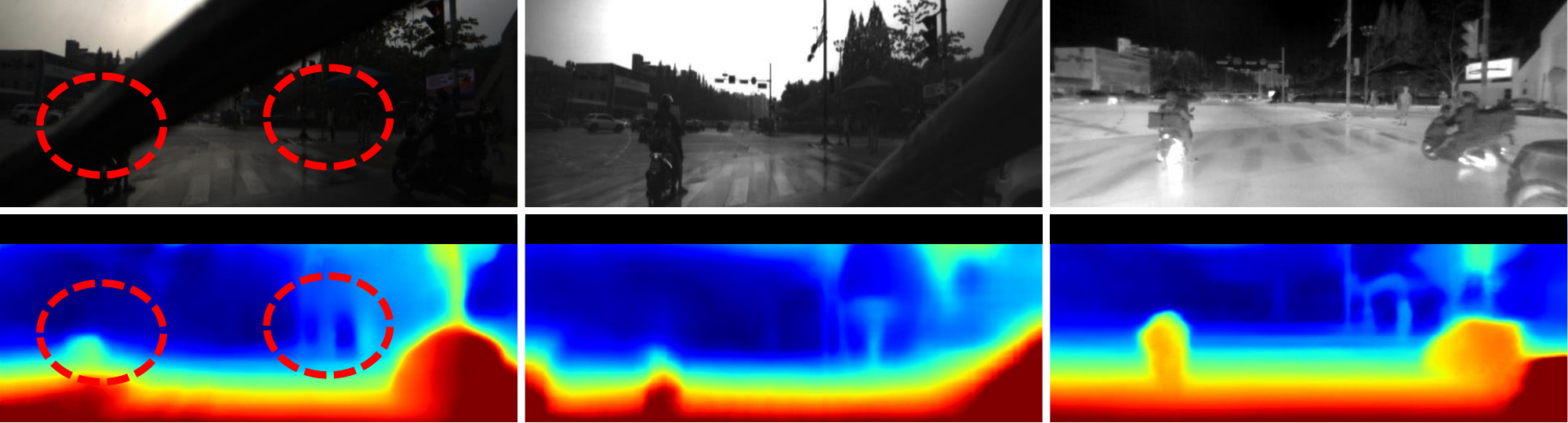} &  \includegraphics[width=0.49\linewidth]{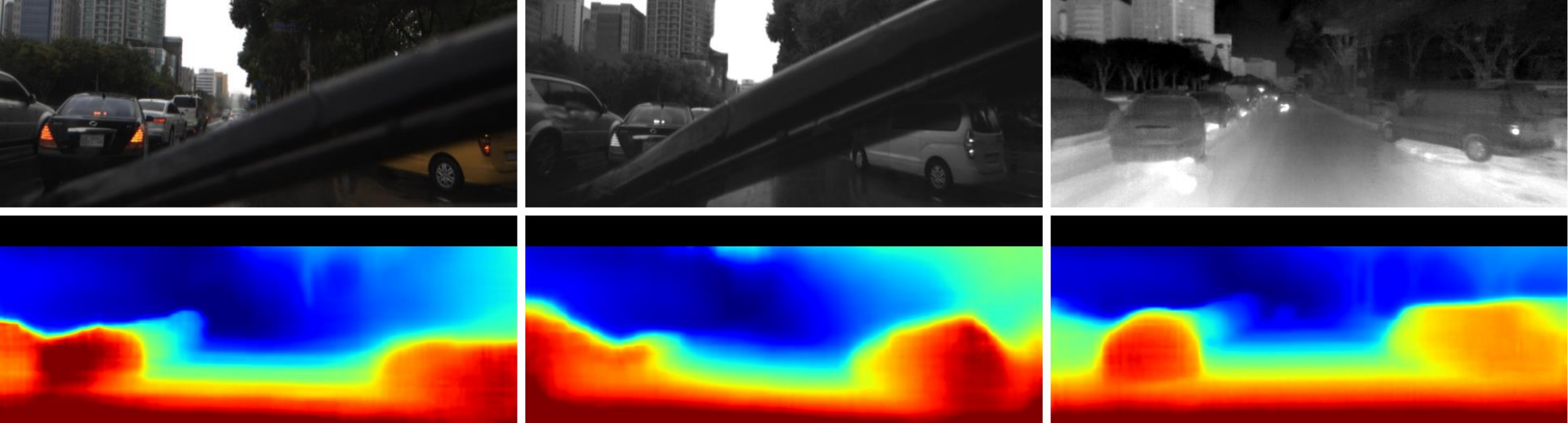} \vspace{-0.02in} \\ 
\multicolumn{2}{c}{{\footnotesize (c) Depth maps from RGB, NIR, and thermal images (MS$^2$ Depth - Rainy)}} \\
\end{tabular}
}
\end{center}
\vspace{-0.1in}
\caption{{\bf Monocular depth map comparison between RGB, NIR, and thermal images on MS$^2$ depth dataset}. 
The figure illustrates monocular depth maps estimated from RGB, NIR, and thermal images under (a) daytime, (b) nighttime, and (c) rainy conditions. 
Each row consists of input images (top) and corresponding depth maps (bottom). 
Red circles highlight key regions where depth estimation varies across sensor modalities, revealing challenges in handling low-light conditions and occlusions.
}
\label{fig:ms2_bench_mono}
\vspace{-0.2in}
\end{figure*}

\subsection{Monocular Depth Estimation from RGB, NIR, and Thermal}
\label{subsec:exp_result}
{\bf Monocular depth estimation.} 
Monocular Depth Estimation (MDE) aims to predict a depth map $\mathbf{D} \in \mathbb{R}^{H \times W}$ from a monocular image $\mathbf{I} \in \mathbb{R}^{H \times W \times C_{\text{in}}}$ using a neural network $\psi_{\theta}$.
The inference process can be formalized as follows:
\begin{equation}
    \mathbf{D} = \psi_{\theta}(\mathbf{I}).
\end{equation}
where $\mathbf{I}$ and $\psi_{\theta}$ can vary depending on modality. 
RGB images have $C_{\text{in}}=3$, while NIR and thermal images have $C_{\text{in}}=1$.
To enable seamless reuses of pre-built RGB-based neural architecture for NIR and thermal image inputs, we repeat their single-channel data three times along the channel axis, augmenting to $C_{\text{in}}=3$.
Also, we train networks $\psi_{\theta}$ in modality-wise manner (\ie, $\psi_{\theta}^{RGB}$, $\psi_{\theta}^{NIR}$, $\psi_{\theta}^{THR}$).

{\bf Evaluation protocol.} 
This section provides benchmarking results of representative MDE models (\ie, DORN, BTS, AdaBins, and NeWCRF) by training them on RGB, NIR, and thermal modalities\footnote{Note that each model is trained with its corresponding loss function as specified in its official source code.} and testing them on MS$^2$ day, night, and rainy test sequences.
For a fair comparison, all modalities have the same resolutions of 640$\times$256.
The evaluation metrics for depth estimations follow widely used standard metrics, such as Absolute Relative Error (AbsRel), Squared Relative Error (SqRel), Root Mean Square Error (RMSE), and RMSE logarithm (RMSElog), along with threshold-based accuracy metrics ($\delta < 1.25, \delta  < 1.25^2, \delta  < 1.25^3$). 
Please refer to \cite{eigen2014depth} for detailed explanations and equations.


{\bf Findings 1. Modern architecture significantly impacts performance across modalities.} 
As shown in~\cref{table:ms2_bench_mono}, NeWCRF consistently outperforms other models (\ie, DORN, BTS, AdaBins) across all modalities and conditions, achieving the lowest RMSE and highest $\delta$ accuracy. 
Unlike BTS and DORN, which rely on CNNs, NeWCRF employs a transformer architecture~\cite{liu2021swin} in both backbone and decoder modules, allowing it to effectively aggregate non-local and global feature information with an attention mechanism.
This demonstrates that advanced transformer architectures are also effective for non-conventional sensor modalities (\ie, NIR and thermal) and challenging environments (\ie, night and rainy).

{\bf Findings 2. NIR demonstrates better depth estimation results than RGB images at nighttime.} 
NIR cameras are designed to capture the near-infrared spectrum (700-1,000nm) beyond the visible light spectrum (400-700nm), allowing them to provide clean images in low-light conditions due to higher sensor sensitivity, reduced atmospheric scattering, and ambient infrared radiation.
Consequently, depth estimation at night is more accurate with NIR images than with RGB images, as shown in \cref{table:ms2_bench_mono}-(a,b) and red circles in \cref{fig:ms2_bench_mono}-(b). 
The tendency holds across all models. Notably, on the night test set, NeWCRF achieves RMSE of 3.157 with NIR images, which is better than 3.573 with RGB images.
However, this high sensor sensitivity often causes saturation effects on reflective surfaces by car headlights, reducing reliability in certain regions.

{\bf Findings 3. Thermal image provides the most robust depth estimation across all conditions.} 
Depth maps estimated from thermal images consistently outperform those from RGB and NIR across daytime, nighttime, and rainy conditions. 
This tendency holds across all tested models, including DORN, BTS, AdaBins, and NeWCRF. 
Specifically, NeWCRF achieves an average RMSE of 3.729 (RGB), 3.791 (NIR), and 2.937 (thermal). 
The performance gap is particularly noticeable in challenging environments. 
While thermal modality achieves 0.3-0.4 lower RMSE values at daytime (\ie, RGB: 3.111, NIR: 3.071, Thermal: 2.717), the difference grows significantly at night (\ie, RGB: 3.573, NIR: 3.157, Thermal: 2.544) and in rainy conditions (\ie, RGB: 4.447, NIR: 5.042, Thermal: 3.503).

This superior performance is attributed to the thermal camera property that is unaffected by lighting changes and reflections, ensuring its effectiveness in extreme conditions.
For instance, thermal image shows better object distinguishability compared to the other modalities, such as higher contrast between a bus and motorcycle (\ie, \cref{fig:ms2_bench_mono}-(a)) and between a wall and parked car (\cref{fig:ms2_bench_mono}-(b)).
In rainy conditions, thermal image is less affected by reflections, glare, occlusion, and water-induced distortions that degrade RGB-based depth estimation (\ie, \cref{fig:ms2_bench_mono}-(c)). 
These advantages make thermal imaging a robust alternative for depth perception in diverse and adverse environments.

\begin{figure*}[t!]
\begin{center}
{
\begin{tabular}{c@{\hskip 0.01\linewidth}c@{\hskip 0.01\linewidth}c}
\includegraphics[width=0.33\linewidth]{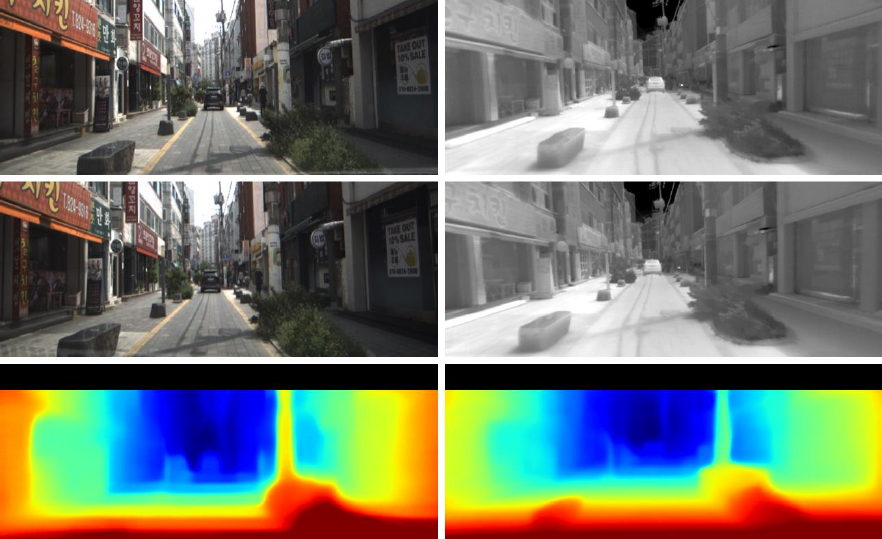} &  \includegraphics[width=0.33\linewidth]{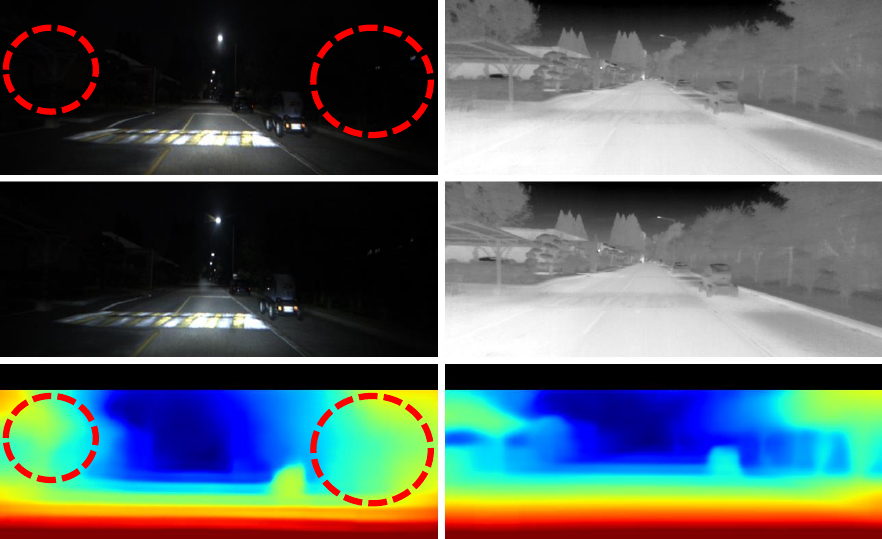} & 
\includegraphics[width=0.33\linewidth]{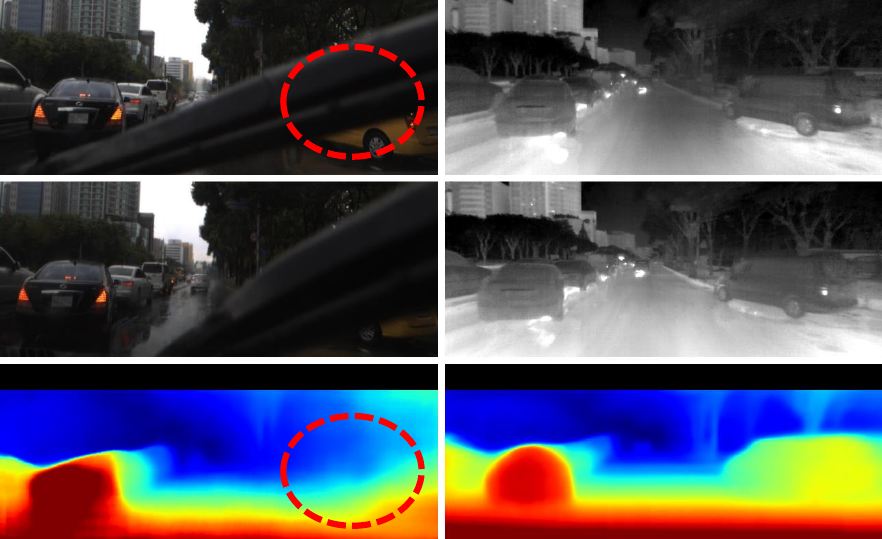} \vspace{-0.02in} \\  
{\footnotesize (a) MS$^2$ Depth - Daytime} & {\footnotesize (b) MS$^2$ Depth - Nightime} & {\footnotesize (c) MS$^2$ Depth - Rainy} \\
\includegraphics[width=0.33\linewidth]{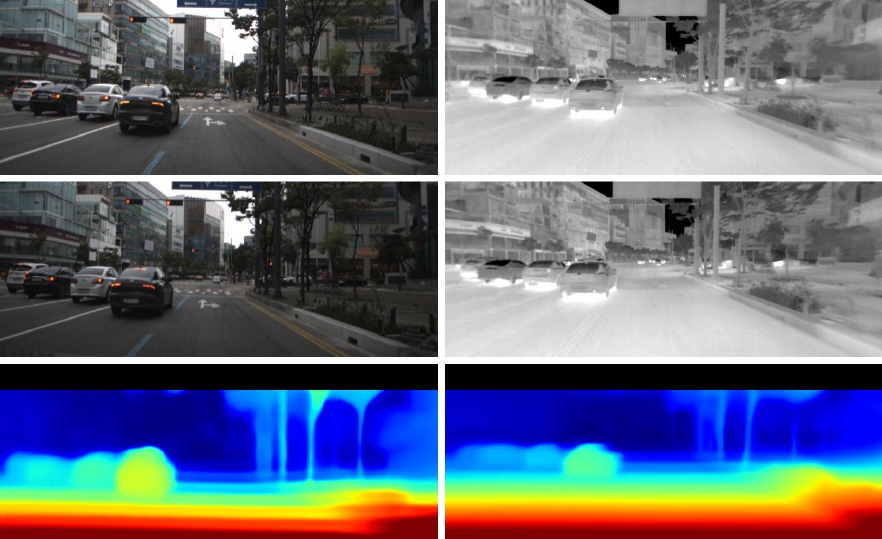} & 
\includegraphics[width=0.33\linewidth]{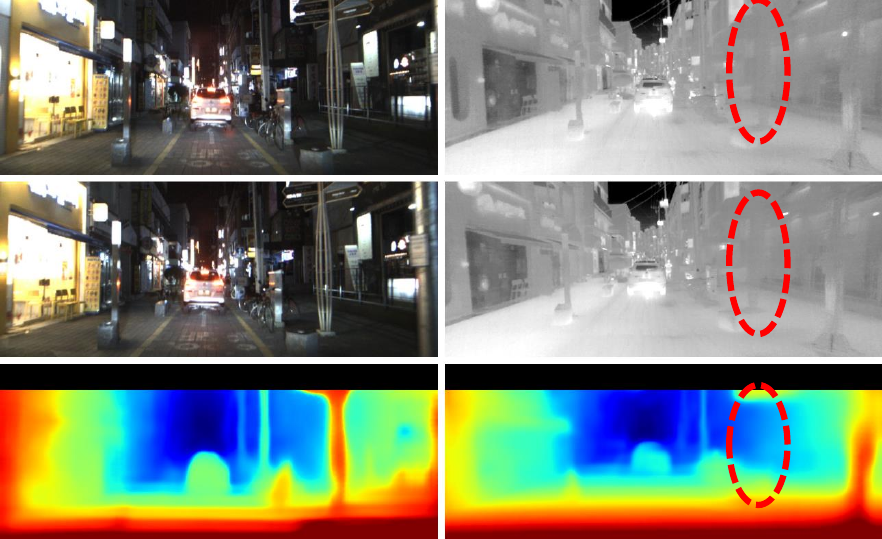} &  \includegraphics[width=0.33\linewidth]{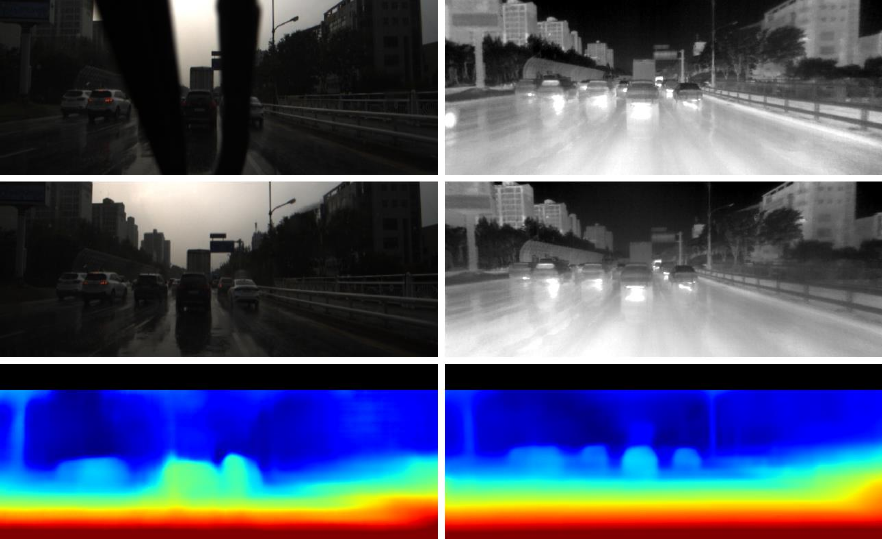} \vspace{-0.02in} \\ 
{\footnotesize (d) MS$^2$ Depth - Daytime} & {\footnotesize (e) MS$^2$ Depth - Nightime} & {\footnotesize (f) MS$^2$ Depth - Rainy} \\ 
\end{tabular}
}
\end{center}
\vspace{-0.1in}
\caption{{\bf Disparity map comparison between RGB and thermal images on MS$^2$ depth dataset}. 
The figure illustrates disparity maps estimated from RGB and thermal images under (a,d) daytime, (b,e) nighttime, and (c,f) rainy conditions. 
Each row consists of input stereo images (first, second) and corresponding disparity maps (third). 
Red circles highlight key regions where disparity estimation varies across sensor modalities, revealing challenges in handling low-light conditions, insufficient texture, and occlusions.}
\label{fig:ms2_bench_stereo}
\vspace{-0.2in}
\end{figure*}

{\bf Findings 4. Rainy conditions are the most challenging across all modalities.}
All modalities exhibit the highest errors in rainy conditions. 
For RGB and NIR spectrums, the performance degradation is mainly caused by light scattering, energy absorption, light reflection, and occlusion by water or raindrops, along with occlusion by windshield wipers.
These factors produce low-contrast, blurry, and missing image details, making inaccurate depth estimation results, as shown in \cref{fig:ms2_bench_mono}-(c).
In particular, for the NIR spectrum, the energy absorption ratio by water is higher than the visible spectrum~\cite{hale1973optical}, leading to generally worse RMSE results than the RGB modality.
In contrast, thermal images remain the most robust performance compared to other modalities because the thermal infrared spectrum is less affected by water particles.
However, water absorption still reduces thermal radiation, leading to lower contrast and increased noise compared to daytime thermal images.
Additionally, the thermal reflection by wet surfaces is another problem in thermal images.


\begin{table*}[t!]
\caption{\textbf{Benchmarking results of SDE networks on the MS$^2$ depth dataset across RGB and thermal modalities}.
The table presents benchmarking results of representative SDE networks~\cite{chang2018pyramid,guo2019group,xu2020aanet,xu2022attention} with RGB and thermal modalities on MS$^2$ depth evaluation sets (\ie, daytime, nighttime, and rainy).
The best performance in each modality is highlighted in \textbf{bold}. 
}
\begin{center}
\subfloat[Disparity estimation results on the MS$^2$ depth evaluation sets (Left: RGB, Right: thermal).]
{ 
    \resizebox{0.9\linewidth}{!}{
    \def\arraystretch{1.0}
    \footnotesize
    \begin{tabular}{|c|c|c|c|cc||c|c|cc|}
    \hline
    \multirow{2}{*}{Methods} & \multirow{2}{*}{TestSet} & \multicolumn{4}{c||}{RGB} & \multicolumn{4}{c|}{Thermal} \\ \cline{3-10}
     &  &  EPE-all(px) & D1-all(\%) & $>1$px(\%) & $>2$px(\%) &  EPE-all(px) & D1-all(\%) & $>1$px(\%) & $>2$px(\%) \\ 
    \hline \hline
    \multirow{4}{*}{PSMNet~\cite{chang2018pyramid}} 
    & Day   & 0.244 & 0.385 & 2.432 & 0.764 & 0.253 & 0.266 & 2.372 & 0.582 \\
    & Night & 0.282 & 0.270 & 2.819 & 0.601 & 0.331 & 0.286 & 5.512 & 1.018 \\
    & Rain  & 0.654 & 2.187 & 17.522 & 4.935 & 0.294 & 0.221 & 3.545 & 0.625 \\ \cline{2-10}
    & \cellcolor{Gray1}Avg   & \cellcolor{Gray1}0.400 & \cellcolor{Gray1}\textbf{0.981} & \cellcolor{Gray1}7.856 & \cellcolor{Gray1}\textbf{2.177} & \cellcolor{Gray1}0.292 & \cellcolor{Gray1}0.257 & \cellcolor{Gray1}3.794 & \cellcolor{Gray1}0.737\\ \hline
    \multirow{4}{*}{GwcNet~\cite{guo2019group}} 
    & Day   & 0.234 & 0.406 & 2.160 & 0.734 & 0.246 & 0.241 & 2.278 & 0.569 \\
    & Night & 0.263 & 0.228 & 2.393 & 0.525 & 0.324 & 0.243 & 5.208 & 0.947 \\
    & Rain  & 0.684 & 2.487 & 18.751 & 5.789 & 0.285 & 0.187 & 3.261 & 0.552 \\ \cline{2-10}
    & \cellcolor{Gray1}Avg   & \cellcolor{Gray1}0.401 & \cellcolor{Gray1}1.080 & \cellcolor{Gray1}8.062 & \cellcolor{Gray1}2.442 & \cellcolor{Gray1}0.285 & \cellcolor{Gray1}\textbf{0.223} & \cellcolor{Gray1}\textbf{3.565} & \cellcolor{Gray1}\textbf{0.685}\\ \hline
    \multirow{4}{*}{AANet~\cite{xu2020aanet}} 
    & Day   & 0.213 & 0.345 & 2.076 & 0.675 & 0.240 & 0.336 & 2.374 & 0.678 \\
    & Night & 0.262 & 0.259 & 2.762 & 0.572 & 0.327 & 0.314 & 5.405 & 1.027 \\
    & Rain  & 0.697 & 2.661 & 18.961 & 6.088 & 0.285 & 0.267 & 3.601 & 0.737 \\ \cline{2-10}
    & \cellcolor{Gray1}Avg   & \cellcolor{Gray1}0.399 & \cellcolor{Gray1}1.131 & \cellcolor{Gray1}8.227 & \cellcolor{Gray1}2.543 & \cellcolor{Gray1}0.284 & \cellcolor{Gray1}0.305 & \cellcolor{Gray1}3.780 & \cellcolor{Gray1}0.811 \\ \hline
    \multirow{4}{*}{ACVNet~\cite{xu2022attention}} 
    & Day   & 0.233 & 0.401 & 2.202 & 0.736 & 0.241 & 0.322 & 2.324 & 0.652 \\
    & Night & 0.254 & 0.221 & 2.375 & 0.510 & 0.313 & 0.244 & 4.969 & 0.895 \\
    & Rain  & 0.669 & 2.341 & 17.904 & 5.546 & 0.283 & 0.195 & 3.442 & 0.591 \\ \cline{2-7}
    & \cellcolor{Gray1}Avg   & \cellcolor{Gray1}\textbf{0.393} & \cellcolor{Gray1}{1.024} & \cellcolor{Gray1}\textbf{7.772} & \cellcolor{Gray1}2.353 & \cellcolor{Gray1}\textbf{0.279} & \cellcolor{Gray1}0.252 & \cellcolor{Gray1}3.567 & \cellcolor{Gray1}0.709 \\ \hline
    \end{tabular}
    }
}
\vspace{3mm}
\subfloat[Stereo depth results on the MS$^2$ depth evaluation sets (\textbf{\textit{Modality: RGB images}}).]
{ 
    \resizebox{0.9 \textwidth}{!}{
    \def\arraystretch{0.9}
    \footnotesize
    \begin{tabular}{|c|c|cccc|ccc|}
    \hline
    \multirow{2}{*}{Methods} & \multirow{2}{*}{TestSet} & \multicolumn{4}{c|}{\textbf{Error $\downarrow$}} & 
    \multicolumn{3}{c|}{\textbf{Accuracy $\uparrow$}}    \\ \cline{3-9}
     &  &  AbsRel & SqRel & RMSE & RMSElog &  $\delta < 1.25$ & $\delta < 1.25^{2}$ & $\delta < 1.25^{3}$ \\ 
    \hline \hline
    \multirow{4}{*}{PSMNet~\cite{chang2018pyramid}} 
    & Day   & 0.033 & 0.085 & 1.664 & 0.056 & 0.992 & 0.998 & 0.999 \\
    & Night & 0.040 & 0.099 & 1.820 & 0.060 & 0.991 & 0.999 & 1.000 \\
    & Rain  & 0.091 & 0.554 & 3.892 & 0.139 & 0.921 & 0.981 & 0.992 \\ \cline{2-9}
    & \cellcolor{Gray1}Avg   & \cellcolor{Gray1}0.056 & \cellcolor{Gray1}0.254 & \cellcolor{Gray1}2.497 & \cellcolor{Gray1}0.086 & \cellcolor{Gray1}0.967 & \cellcolor{Gray1}\textbf{0.993} & \cellcolor{Gray1}\textbf{0.997} \\ \hline
    \multirow{4}{*}{GwcNet~\cite{guo2019group}} 
    & Day   & 0.033 & 0.088 & 1.661 & 0.056 & 0.992 & 0.998 & 0.999 \\
    & Night & 0.038 & 0.086 & 1.645 & 0.056 & 0.993 & 0.999 & 1.000 \\
    & Rain  & 0.093 & 0.584 & 4.054 & 0.142 & 0.914 & 0.980 & 0.991 \\ \cline{2-9}
    & \cellcolor{Gray1}Avg   & \cellcolor{Gray1}0.056 & \cellcolor{Gray1}0.261 & \cellcolor{Gray1}2.496 & \cellcolor{Gray1}0.086 & \cellcolor{Gray1}0.965 & \cellcolor{Gray1}0.992 & \cellcolor{Gray1}\textbf{0.997} \\ \hline
    \multirow{4}{*}{AANet~\cite{xu2020aanet}} 
    & Day   & 0.029 & 0.072 & 1.465 & 0.051 & 0.993 & 0.998 & 0.999 \\
    & Night & 0.037 & 0.083 & 1.569 & 0.056 & 0.992 & 0.999 & 1.000 \\
    & Rain  & 0.093 & 0.600 & 4.114 & 0.147 & 0.914 & 0.977 & 0.990 \\ \cline{2-9}
    & \cellcolor{Gray1}Avg   & \cellcolor{Gray1}0.054 & \cellcolor{Gray1}0.261 & \cellcolor{Gray1}2.429 & \cellcolor{Gray1}0.086 & \cellcolor{Gray1}0.965 & \cellcolor{Gray1}0.991 & \cellcolor{Gray1}0.996 \\ \hline
    \multirow{4}{*}{ACVNet~\cite{xu2022attention}} 
    & Day   & 0.032 & 0.085 & 1.585 & 0.054 & 0.992 & 0.998 & 0.999 \\
    & Night & 0.035 & 0.072 & 1.485 & 0.052 & 0.995 & 0.999 & 1.000 \\
    & Rain  & 0.091 & 0.569 & 3.983 & 0.140 & 0.920 & 0.981 & 0.992 \\ \cline{2-9}
    & \cellcolor{Gray1}Avg   & \cellcolor{Gray1}\textbf{0.054} & \cellcolor{Gray1}\textbf{0.251} & \cellcolor{Gray1}\textbf{2.395} & \cellcolor{Gray1}\textbf{0.083} & \cellcolor{Gray1}\textbf{0.968} & \cellcolor{Gray1}\textbf{0.993} & \cellcolor{Gray1}\textbf{0.997} \\ \hline
    \end{tabular}
    }
}
\vspace{3mm}
\subfloat[Stereo depth estimation results on the MS$^2$ depth evaluation sets (\textbf{\textit{Modality: thermal images}}).]
{
    \resizebox{0.9 \textwidth}{!}{
    \def\arraystretch{0.9}
    \footnotesize
    \begin{tabular}{|c|c|cccc|ccc|}
    \hline
    \multirow{2}{*}{Methods} & \multirow{2}{*}{TestSet} & \multicolumn{4}{c|}{\textbf{Error $\downarrow$}} & 
    \multicolumn{3}{c|}{\textbf{Accuracy $\uparrow$}}    \\ \cline{3-9}
     &  &  AbsRel & SqRel & RMSE & RMSElog &  $\delta < 1.25$ & $\delta < 1.25^{2}$ & $\delta < 1.25^{3}$ \\ 
    \hline \hline
    \multirow{4}{*}{PSMNet~\cite{chang2018pyramid}} 
    & Day   & 0.033 & 0.070 & 1.533 & 0.051 & 0.995 & 0.999 & 1.000 \\
    & Night & 0.042 & 0.089 & 1.617 & 0.060 & 0.991 & 1.000 & 1.000 \\
    & Rain  & 0.040 & 0.098 & 1.802 & 0.059 & 0.993 & 0.999 & 1.000 \\ \cline{2-9}
    & \cellcolor{Gray1}Avg   & \cellcolor{Gray1}0.038 & \cellcolor{Gray1}0.086 & \cellcolor{Gray1}1.654 & \cellcolor{Gray1}0.057 & \cellcolor{Gray1}0.993 & \cellcolor{Gray1}\textbf{0.999} & \cellcolor{Gray1}\textbf{1.000} \\ \hline
    \multirow{4}{*}{GwcNet~\cite{guo2019group}} 
    & Day   & 0.032 & 0.067 & 1.513 & 0.049 & 0.995 & 0.999 & 1.000 \\
    & Night & 0.042 & 0.086 & 1.580 & 0.059 & 0.992 & 1.000 & 1.000 \\
    & Rain  & 0.039 & 0.091 & 1.721 & 0.057 & 0.994 & 0.999 & 1.000 \\ \cline{2-9}
    & \cellcolor{Gray1}Avg   & \cellcolor{Gray1}0.038 & \cellcolor{Gray1}0.082 & \cellcolor{Gray1}1.608 & \cellcolor{Gray1}0.055 & \cellcolor{Gray1}\textbf{0.994} & \cellcolor{Gray1}\textbf{0.999} & \cellcolor{Gray1}\textbf{1.000} \\ \hline
    \multirow{4}{*}{AANet~\cite{xu2020aanet}} 
    & Day   & 0.030 & 0.050 & 1.203 & 0.046 & 0.996 & 0.999 & 1.000 \\
    & Night & 0.041 & 0.079 & 1.442 & 0.058 & 0.992 & 1.000 & 1.000 \\
    & Rain  & 0.037 & 0.079 & 1.532 & 0.055 & 0.994 & 0.999 & 1.000 \\ \cline{2-9}
    & \cellcolor{Gray1}Avg   & \cellcolor{Gray1}\textbf{0.036} & \cellcolor{Gray1}\textbf{0.070} & \cellcolor{Gray1}\textbf{1.395} & \cellcolor{Gray1}\textbf{0.053} & \cellcolor{Gray1}\textbf{0.994} & \cellcolor{Gray1}\textbf{0.999} & \cellcolor{Gray1}\textbf{1.000} \\ \hline
    \multirow{4}{*}{ACVNet~\cite{xu2022attention}} 
    & Day   & 0.031 & 0.065 & 1.465 & 0.049 & 0.995 & 0.999 & 1.000 \\
    & Night & 0.040 & 0.082 & 1.526 & 0.057 & 0.992 & 1.000 & 1.000 \\
    & Rain  & 0.039 & 0.091 & 1.724 & 0.056 & 0.994 & 0.999 & 1.000 \\ \cline{2-9}
    & \cellcolor{Gray1}Avg   & \cellcolor{Gray1}0.037 & \cellcolor{Gray1}0.080 & \cellcolor{Gray1}{1.576} & \cellcolor{Gray1}{0.054} & \cellcolor{Gray1}\textbf{0.994} & \cellcolor{Gray1}\textbf{0.999} & \cellcolor{Gray1}\textbf{1.000} \\ \hline
    \end{tabular}
    }    
}
    
\end{center}
\vspace{-0.1in}
\label{table:ms2_bench_stereo}
\end{table*}

\subsection{Stereo Depth Estimation from RGB and Thermal images}
{\bf Disparity estimation.} 
Disparity estimation predicts a disparity map $\mathbf{d} \in \mathbb{R}^{H \times W}$, representing horizontal pixel displacement between corresponding points in a stereo image pair, using a neural network $\psi_{\theta}$ with a given left and right stereo images $\mathbf{I}_L, \mathbf{I}_R \in \mathbb{R}^{H \times W \times C_{\text{in}}}$.
In addition, the disparity $\mathbf{d}$ can be converted to depth $\mathbf{D}$ using stereo camera baseline and focal length.
The inference and conversion processes can be formalized as follows:
\begin{equation}
    \mathbf{d} = \psi_{\theta}(\mathbf{I_L}, \mathbf{I_R}), \quad
    \mathbf{D} = \frac{f\cdot B}{\mathbf{d}},    
\end{equation}
where $f$ is the focal length and $B$ is the stereo camera baseline.
To enable the seamless reuse of disparity networks designed for RGB images, NIR and thermal images are augmented to match input channel dimension $C_{\text{in}}=3$.
Additionally, we train networks $\psi_{\theta}$ per modality.

{\bf Evaluation protocol.} 
This section provides benchmarking results of representative disparity estimation models (\ie, PSMNet, GwcNet, AANet, and ACVNet) by training them on RGB and thermal modalities\footnote{Note that each model is trained with its corresponding loss function as specified in its official source code.} and testing them on MS$^2$ depth test sets (\ie, day, night, and rainy scenes).
For a fair comparison, all modalities have the same resolutions of 640×256.
NIR stereo was excluded as its baseline is too narrow to predict long-range depth estimation in outdoor cases.
We follow the standard disparity evaluation metrics, including End-Point Error (EPE-all) to quantify the mean absolute disparity error in pixel and percentage metric (\ie, D1-all (\%), $>$1px (\%), and $>$2px (\%)) to measure the percentage of pixels with disparity error exceeding specific thresholds.
For detailed explanations and equations, please refer to~\cite{liang2018learning}.


{\bf Findings 5. Stereo depth estimation significantly outperforms monocular depth estimation in all conditions.} 
As shown in \cref{table:ms2_bench_mono} and \cref{table:ms2_bench_stereo}, stereo depth estimation models consistently achieve substantially lower error metrics (\ie, AbsRel, RMSE) and higher accuracy metrics ($\delta < 1.25^n$) compared to monocular models, regardless of input modalities. 
This performance gap is particularly pronounced in challenging conditions such as nighttime, where monocular depth estimation struggles due to the lack of sufficient texture and illumination, as seen in \cref{fig:ms2_bench_mono} and \cref{fig:ms2_bench_stereo}
(\eg, RGB, NeWCRF RMSE: 3.111 (Day) and 3.573 (Night) $\rightarrow$ ACVNet 1.585 (Day), 1.485 (Night)).
This is because stereo methods utilize spatial consistency between two viewpoints to find optimally matched correspondences rather than monocular depth cues (\eg, object size, occlusion, texture gradient, etc), allowing them to maintain more reliable depth estimation even in low-texture and low-light conditions.

\begin{figure}[t]
\begin{center}
{
\begin{tabular}{c@{\hskip 0.005\linewidth}c}
\includegraphics[width=0.48\linewidth]{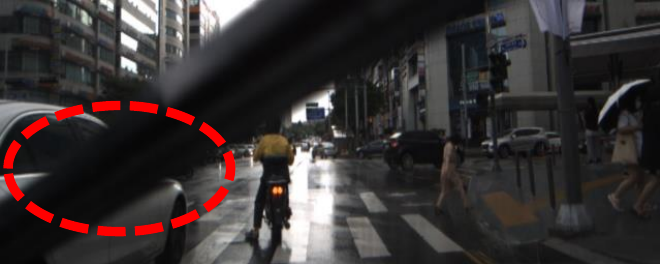} & 
\includegraphics[width=0.48\linewidth]{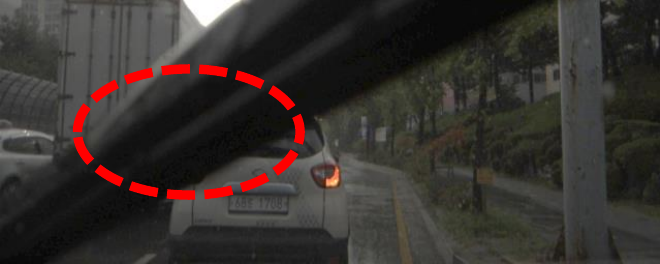} \\  
\multicolumn{2}{c}{{\footnotesize (a) RGB images in MS2-rainy test set}} \\
\includegraphics[width=0.48\linewidth]{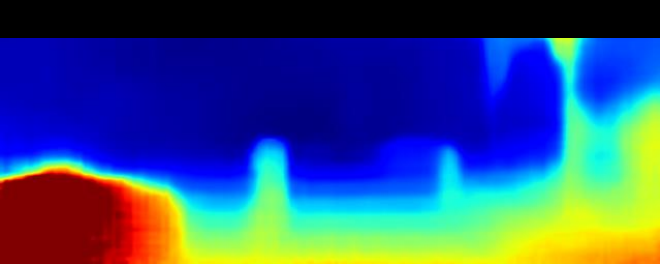} & 
\includegraphics[width=0.48\linewidth]{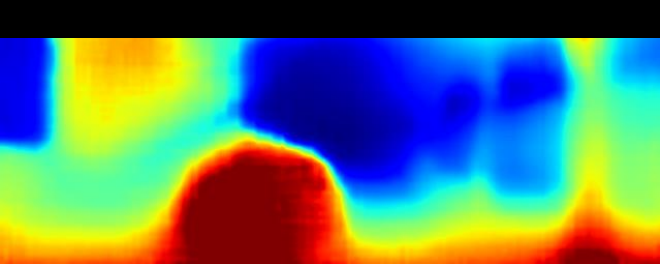} \\  
\multicolumn{2}{c}{{\footnotesize (b) Monocular depth estimation from RGB images}} \\
\includegraphics[width=0.48\linewidth]{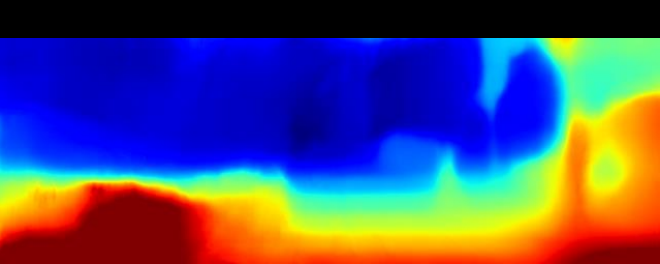} & 
\includegraphics[width=0.48\linewidth]{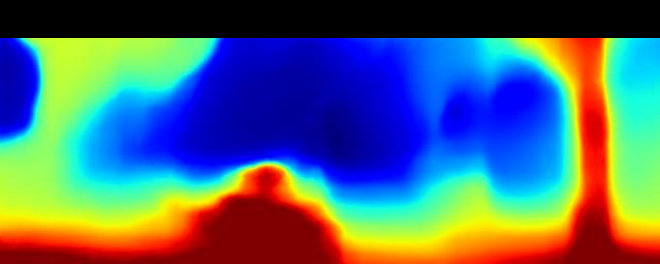} \\  
\multicolumn{2}{c}{{\footnotesize (c) Stereo depth estimation from RGB images}} \\
\end{tabular}
}
\end{center}
\caption{{\bf Depth map comparison in rainy conditions}.
Rain introduces light scattering, windshield wiper occlusions, blurriness, and spatial inconsistencies in RGB images, leading to degraded stereo depth maps. 
In contrast, monocular depth estimation can compensate for occluded regions using non-local features, resulting in more stable depth predictions.
}
\label{fig:challenge_rgb_mono_stereo}
\vspace{-0.10in}
\end{figure}

{\bf Findings 6. RGB stereo matching struggles in rainy conditions.} 
On the other hand, the performance improvement from monocular to stereo models is not significant in RGB modality (\eg, RGB, NeWCRF RMSE: 4.447 (rainy) $\rightarrow$ ACVNet RMSE: 3.983 (rainy)). 
This relatively small improvement suggests that stereo matching in the RGB modality faces inherent challenges in rainy conditions. 
As observed in \cref{fig:ms2_bench_stereo}, rain introduces reflections, water droplets, blur, and windshield wiper operation, which leads to frequent spatial inconsistency between the left and right images, making the stereo matching process further tricky (\eg, RGB, ACVNet EPE: 0.233 (day) vs. 0.669 (rainy)).
Additionally, the reduced contrast and dynamic lighting variations caused by wet road surfaces and vehicle headlights further complicate the stereo matching process. 
Therefore, the depth estimations from monocular and stereo RGB images in rainy conditions are comparable or, in some cases, monocular depth estimation is better than stereo matching, as shown in \cref{fig:challenge_rgb_mono_stereo}.

{\bf Findings 7. Thermal stereo matching achieves the most stable depth estimation across all conditions.} 
As observed in \cref{table:ms2_bench_stereo}-(a), the performance difference between RGB and thermal stereo is not significant in day and night conditions since stereo matching relies on spatial consistency along with distinguishable features (\eg, edge, color, subtle details).
Therefore, the insufficient textures in thermal images are not beneficial for matching problems.
However, after converting disparity to depth, thermal stereo matching consistently maintains lower error and higher accuracy across all conditions, as shown in \cref{fig:ms2_bench_stereo}-(c) and \cref{table:ms2_bench_stereo}-(c).

This is because disparity metrics are highly sensitive to pixel-level errors, especially for nearby objects. 
However, since depth is inversely related to disparity, even small disparity errors in distant objects can lead to significant depth errors, making depth evaluation metrics more influenced by far-distance regions.
Therefore, in general, RGB stereo can provide accurate matching results, especially for close objects, but not for distant objects due to the defocus blur and reduced texture.
In contrast, thermal stereo remains stable across varying distances despite its lack of fine-grained texture.

This stability can be attributed to the thermal modality’s insensitivity to lighting changes, reflections, and motion blur, which are key sources of error in RGB stereo matching.
Furthermore, most nighttime images in the MS$^2$ dataset include ambient lighting from car headlights or streetlights, allowing RGB images to remain recognizable in nighttime scenarios.
However, as shown in \cref{fig:ms2_bench_stereo}-(b), extremely low-light conditions make most content indistinguishable, ultimately leading to mispredictions in RGB stereo.
This confirms that thermal stereo is highly reliable and effective for depth estimation in more extreme environments, such as complete darkness, heavy fog, rainy night, fire, and smoke.

{\bf Findings 8. Thermal stereo matching degrades under severe noise and low contrast in night and rainy conditions.} 
Thermal stereo performances in nighttime and rainy conditions are generally lower than in daytime conditions across all stereo models (\eg, Thermal, AANet RMSE: 1.465 (day), 1.442 (night), 1.532 (rainy)).
This degradation comes from increased sensor noise and reduced thermal contrast. 
Compared to daytime, which also provides thermal reflection by the sun, thermal cameras mostly rely on thermal radiation from objects at night, leading to lower contrast and fewer distinguishable features.
Similarly, in rainy conditions, energy absorption by water particles in the atmosphere or on the lens reduces image contrast and introduces additional noise, as shown in \cref{fig:ms2_bench_stereo}-(c,f) and \cref{fig:challenge_sensor_noise}. 
On top of insufficient texture issues, these factors make accurate pixel-level matching even more challenging.

\section{Discussion \& Challenges}
\label{sec:discussion}
\subsection{Why does monocular depth from thermal image outperform RGB image even in daytime conditions?}
As shown in~\cref{table:ms2_bench_mono}-(a) and (c), monocular depth estimation consistently outperforms in thermal modality even in daytime conditions.
We believe there are two potential reasons: 1) monocular depth cues and 2) thermal image measurement process.
Monocular depth estimation primarily relies on perspective geometry and monocular depth cues derived from a single image, such as object size, occlusion, linear perspective (\eg, line and vanishing point), and texture gradient~\cite{hu2019visualization}.
While these cues are shared across different sensing modalities, RGB images are susceptible to lighting variations, including overexposure, underexposure, shadows, reflections, and sudden lighting changes. 
These factors can disrupt texture gradients, contrast, and color, leading to inconsistent depth predictions. 
In contrast, thermal images are invariant to visible lighting conditions, such as strong sunlight and shadows, preserving monocular depth cues in general.

\begin{figure}[t]
\centering
\resizebox{0.98\columnwidth}{!}
{
\begin{tabular}{c}
\includegraphics[width=0.98\linewidth]{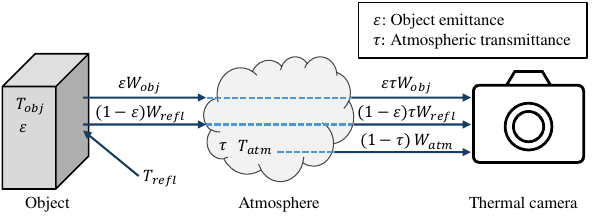} \\
\end{tabular}
}
\caption{{\bf General thermographic measurement schematic}.
Thermal cameras measure an object radiation, which is influenced by emission from the object, reflected emission from ambient sources, and emission from the atmosphere~\cite{flir-ax5}.
}
\label{fig:thermal_measure}
\end{figure}

Additionally, the thermal camera measurement process inherently relates to depth. 
The following equation~\eqref{equ:principal} and~\cref{fig:thermal_measure} are the measurement formula of standard thermal camera~\cite{flir-ax5}, showing the measured radiation of an object depends on its temperature, ambient heat source, and atmosphere between the object and camera.
\begin{equation} 
\label{equ:principal}
W_{total} =\varepsilon\tau W_{obj} + (1-\varepsilon)\tau W_{refl} + (1-\tau)W_{atm} , 
\end{equation}
where $\varepsilon$ is the object emittance, $\tau$ is the atmospheric transmittance, $W_{obj}$ is the thermal radiation emitted by objects, $W_{refl}$ is the reflected emission from ambient sources, and $W_{atm}$ is the emission from the atmosphere.

Assuming a uniform temperature across the scene and minimal surrounding objects, the total measured radiation $W_{total}$ is primarily influenced by $W_{obj}$ and $W_{atm}$. 
Since $W_{atm}$ remains nearly constant, the object's measured radiation $W_{obj}$ is inversely proportional to the distance $D$ (\ie, $W_{obj} \varpropto T_{obj}/D$). 
This suggests that, under uniform temperature conditions (\eg, road), thermal radiation can encode depth information.

\subsection{Robust 3D Geometry from Thermal Images}
Autonomous driving systems rely on various 3D geometry tasks to perceive and understand their surroundings. 
These tasks include depth estimation~\cite{yang2024depth}, optical flow~\cite{teed2020raft}, SLAM~\cite{ qin2018vins}, 3D reconstruction~\cite{agarwal2011building}, occupancy~\cite{zhang2023occformer}, NeRF~\cite{mildenhall2021nerf}, 3D object detection~\cite{brazil2023omni3d}, and 6D pose estimation~\cite{lin2024sam}.
This manuscript demonstrates the robustness of thermal imaging under adverse weather and lighting conditions, highlighting its potential for reliable and robust 3D perception. 
However, our current work mainly focuses on monocular and stereo depth estimation.
Although recent studies have explored the use of thermal images in NeRF~\cite{ye2024thermal, lu2024thermalgaussian}, depth estimation~\cite{shin2021self, zuo2025monother}, and SLAM~\cite{shin2019sparse}, many research areas remain underexplored. 
Further investigation is needed to achieve comprehensive and robust 3D perception in challenging conditions, enabling higher levels of autonomy.

\subsection{Foundation Model for Non-Conventional Sensors}
Recent advancements in foundation models for vision tasks, such as detection~\cite{liu2024grounding}, segmentation~\cite{kirillov2023segment}, depth~\cite{yang2024depth}, vision-language grounding~\cite{radford2021learning}, have predominantly relied on large-scale RGB dataset. 
However, non-conventional sensors such as thermal, NIR, and SWIR inherently suffer from data and label scarcity problems, making it challenging to develop foundation models from scratch. 
To bridge this gap, a more practical approach is to adapt pre-trained RGB foundation models for non-conventional sensors. 
Given the data and label limitations, this can be achievable by utilizing techniques, such as fine-tuning~\cite{zhang2024irsam}, parameter efficient adaptation~\cite{xiao2024segment}, self-supervised learning~\cite{shin2022maximize}, and contrastive learning~\cite{shin2025bridge}, enabling the development of both sensor-specific or sensor-agnostic foundation models.

\subsection{Multi-Sensor Dataset in Extreme Conditions}
The long-standing goal of robotics and autonomous driving is robust and reliable autonomy under extreme weather and lighting conditions, including heavy rain, smoke, fire, and sand storms.
However, despite the growing interest in robust perception, the lack of large-scale multi-sensor datasets remains a bottleneck for advancing research in extreme conditions. 
A few recent datasets~\cite{zhao2023subtmrs,dhrafani2025firestereo,shin2023deep} provide valuable sensor diversity and benchmarks but still have limitations in terms of coverage across weather, lighting, sensor, and label diversity. 
Future datasets should incorporate more diverse weather conditions (\eg, smoke, fog, sandstorm, heavy rain, snow, and extreme heat), lighting (\eg, complete darkness, overexposure, and varying light), platforms (\eg, vehicle, drone, and mobile robot), seasons (\eg, spring, summer, fall, and winter), sensors (\eg,  event camera, SWIR, 3D/4D mmWave Radar, and spinning/solid-state LiDAR), and labels (\eg, depth, optical flow, segmentation, and 3D detection) for robust perception research and comprehensive testbed.

\begin{figure}[t]
\centering
\resizebox{1.00\columnwidth}{!}
{
\begin{tabular}{c}
\includegraphics[width=1.00\linewidth]{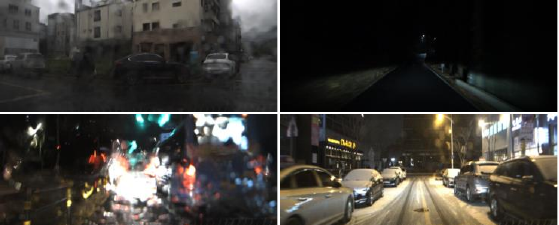} \\
\end{tabular}
}
\caption{{\bf Adverse weather and lighting conditions}.
Further multi-sensor data collection under challenging conditions is essential for exploring the feasibility of high-level autonomy.
}
\vspace{-0.1in}
\label{fig:adverse_weather}
\end{figure}


\subsection{Unifying Monocular and Stereo Depth Cues}
As observed in \cref{sec:experiments}, each monocular and stereo depth estimation has advantages in both RGB and thermal modalities.
Especially for RGB images, monocular depth estimation shows robustness against occlusions from windshield wipers, as shown in \cref{fig:challenge_rgb_mono_stereo}.
For thermal images, stereo depth estimation struggles with low contrast and severe noise, while monocular depth estimation leverages stable monocular cues and thermal-depth relationships, as discussed in \cref{sec:discussion}-A.
Furthermore, although monocular depth estimation is inherently scale-ambiguous, its handiness on a single image enhances robustness against sudden sensor issues (\eg, synchronization delays or sensor disconnections) and spatial inconsistencies caused by occlusion and light scattering.
Therefore, unifying monocular and stereo depth estimation~\cite{zhou2023two,xu2023unifying} can introduce scale awareness while improving overall robustness in challenging conditions. 

\subsection{Robust Correspondence Matching in Thermal Images}
\begin{figure}[t]
\begin{center}
{
\begin{tabular}{c@{\hskip 0.005\linewidth}c}
\includegraphics[width=0.48\linewidth]{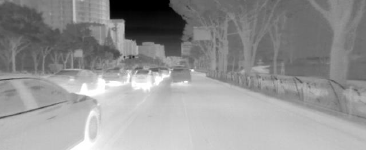} & 
\includegraphics[width=0.48\linewidth]{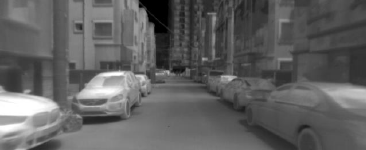} \\  
\multicolumn{2}{c}{{\footnotesize (a) Insufficient textures}} \\
\includegraphics[width=0.48\linewidth]{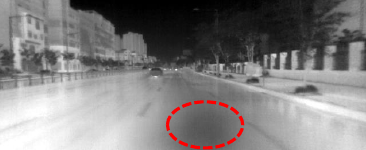} & 
\includegraphics[width=0.48\linewidth]{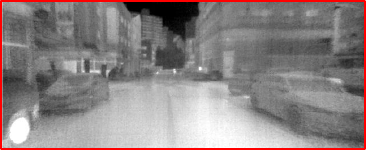} \\
{\footnotesize (b) Water spot on lens} & {\footnotesize (c) Severe noise} \\
\end{tabular}
}
\end{center}
\caption{{\bf Challenges in thermal images}.
Thermal images often suffer from undesirable effects, including insufficient textures, low contrast, and severe noise. They make correspondence matching challenging and lead to depth estimation errors.
}
\label{fig:challenge_sensor_noise}
\vspace{-0.10in}
\end{figure}

Thermal stereo matching often struggles with finding accurate correspondences due to the lack of texture, low contrast, and severe noise, as illustrated in \cref{fig:challenge_sensor_noise}. 
These limitations also affect related tasks such as optical flow, multi-view stereo, feature tracking, and odometry/SLAM, which depend on rich visual textures for reliable matching.
Multiple research directions can be incorporated to address this issue, such as pre-processing techniques that improve contrast~\cite{gil2024fieldscale} and remove fixed pattern noise~\cite{barral2024fixed,saragadam2021thermal} and advanced matching strategies that leverage global context~\cite{xia2022locality}. 
Lastly, designing a corresponding matching method tailored to thermal images could further improve robustness and matching accuracy.

\subsection{Adaptive and Selective Multi-sensor Fusion}
Multi-sensor fusion remains one of the most promising yet challenging areas in perception research. 
While RGB, NIR, thermal, and LiDAR each provide complementary information, their fusion is hindered by spatial misalignment, resolution differences, sensor-specific vulnerability, and domain shift across varying weather and lighting conditions. 
Assessing sensor reliability in adverse environments is particularly challenging, as each modality can fail under specific conditions, such as RGB struggles in darkness and LiDAR in heavy rain.
Further complications arise from modality dropouts, such as sensor disconnection or large occlusion.
Therefore, future research should focus on selective and adaptive sensor fusion strategies that dynamically adjust sensor contributions based on surrounding environmental conditions. 
Some research directions, such as latent space alignment, confidence estimation~\cite{poggi2020uncertainty}, and cross-modal consistency learning~\cite{chen2019crdoco}, can be used to assess sensor reliability by leveraging shared latent space and uncertainty between sensors.
Additionally, RGB-NIR fusion~\cite{park2022adaptive} provides a cost-effective solution for reliable depth estimation in low-light environments, while thermal-LiDAR fusion combines thermal robustness with LiDAR’s scale-awareness, improving depth perception across diverse conditions.

\subsection{Domain Shift in Thermal Images}
Thermal images exhibit lighting-agnostic properties and robustness to various weather conditions.
However, the reliance on thermal radiation could introduce domain shifts influenced by variations in ambient temperature, object emissivity, weather, and seasonal changes. 
Unlike the well-studied domain shift issues in RGB images, the domain shift problem in thermal images remains underexplored.
To address these challenges, research should focus on evaluating domain generalization performance in thermal images across different scenarios, such as outdoor vs. indoor environments~\cite{shin2023deep, lee2022vivid++}, normal vs. smoke-filled conditions~\cite{dhrafani2025firestereo}, and diverse surrounding locations~\cite{zhao2023subtmrs}.
Additionally, developing domain adaptation frameworks specifically tailored for thermal images could improve robustness by learning domain-invariant features that remain stable despite environmental variations. 
\section{Conclusion}
In this study, we introduced the Multi-Spectral Stereo (MS$^2$) dataset, a large-scale 3D vision benchmark that includes stereo RGB, NIR, thermal images, LiDAR data, and GNSS/IMU information.
Our comprehensive evaluation of monocular and stereo depth estimation models across RGB, NIR, and thermal modalities highlights the robustness of thermal imaging, particularly in low-light and adverse weather conditions. 
Furthermore, our analysis reveals numerous findings and interesting research directions, such as domain adaptation on thermal modality, selective sensor fusion, unifying monocular and stereo depth cue, robust correspondence matching, and foundation model for non-conventional sensors, to ensure robust perception in extreme environments.
We hope the dataset and benchmark will serve as a valuable resource for advancing robust perception in autonomous navigation and other real-world robotic applications.

{\small
\bibliographystyle{IEEEtran}
\bibliography{egbib}
}

\vspace{-0.1in}
\begin{IEEEbiography}[{\includegraphics[width=1in,height=1.25in,clip,keepaspectratio]{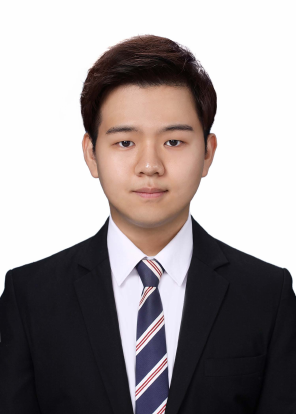}}]{Ukcheol Shin}
is a postdoctoral researcher at Carnegie Mellon University (CMU), affiliated with Robotics Institute. He received his B.S. degree in electrical and information engineering from Seoul National University of Science and Technology, South Korea, in 2017, and his M.S and Ph.D degrees in Electrical Engineering from Korea Advanced Institute of Science and Technology (KAIST), South Korea, in 2019 and 2023, respectively. His research interests include 3D geometry, self-supervised learning, multi-modal sensor fusion, and robot learning. He received the best student paper award in WACV 2023.
\end{IEEEbiography}
\vspace{-0.1in}
\begin{IEEEbiography}[{\includegraphics[width=1in,height=1.25in,clip,keepaspectratio]{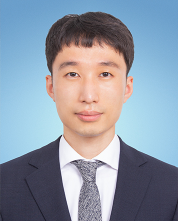}}]{Jinsun Park}
is an assistant professor at the School of Computer Science and Engineering, Pusan National University, Busan, Republic of Korea. From Mar. 2021 to Aug. 2021, he worked as a postdoctoral researcher at Korea Advanced Institute of Science and Technology (KAIST), Daejeon, Republic of Korea. He received his B.S. degree in Electronic Engineering from Hanyang University, Seoul, Republic of Korea, in 2014. He received his M.S. and Ph.D. degrees in Electrical Engineering from KAIST, Daejeon, Republic of Korea, in 2016 and 2021, respectively. From Jul. 2019 to Jan. 2020, he worked as a full-time research intern at HikVision USA, CA, USA.
\end{IEEEbiography}

\end{document}